\documentclass[final]{cvpr}

\usepackage{times}
\usepackage{epsfig}
\usepackage{graphicx}
\usepackage{amsmath}
\usepackage{amssymb}
\usepackage{bbm}
\usepackage{caption}
\usepackage[]{algorithm}
\usepackage{algorithmicx,algpseudocode}
\usepackage{adjustbox}
\usepackage{float}
\usepackage{array,multirow}
\usepackage{booktabs}       % professional-quality tables
\usepackage[pagebackref=true,breaklinks=true,colorlinks,bookmarks=false]{hyperref}

\def\cvprPaperID{3432} % *** Enter the CVPR Paper ID here
\def\confYear{CVPR 2021}

\begin{document}

%%%%%%%%% TITLE
\title{Self-labeled Conditional GANs}

\author{Mehdi Noroozi\\
Bosch Center for Artificial Intelligence\\
{\tt\small mehdi.noroozi@de.bosch.com}
}

\twocolumn[{%
\renewcommand\twocolumn[1][]{#1}%
\maketitle
  \centering
    \vspace{-.15in}
  \includegraphics[width=\linewidth]{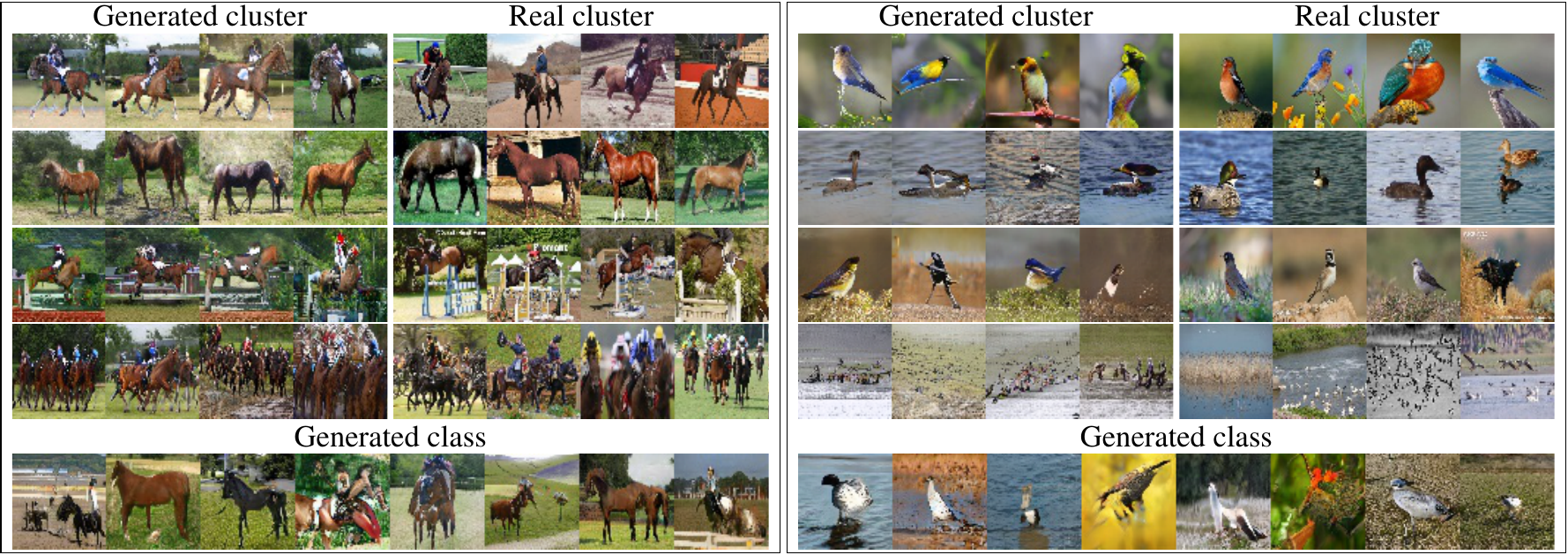}
  \vspace{-.3in}
  \captionof{figure}{%We propose joint training of a clustering network and conditional GANs using no annotation. 
  We jointly train a clustering network that cooperates with the generator to provide the conditional GAN with a set of pseudo-labels.
  We visualize here several clusters dominated by the horse and bird classes for LSUN 20 object categories. Each row shows a cluster and the generated images conditioned on the same pseudo-label. The last row shows the generated images of the conditional GAN trained on ground truth class labels. Our method is trained with $1000$ clusters.
  }
  \vspace{.1in}\vspace{1.5em}
  \label{fig:teaser}
  \vspace{-.1cm}
}]
\maketitle

%%%%%%%%% ABSTRACT
\begin{abstract}
\vspace{-.1cm}
This paper introduces a novel and fully unsupervised framework for conditional GAN training in which labels are automatically obtained from data. We incorporate a clustering network into the standard conditional GAN framework that plays against the discriminator. With the generator, it aims to find a shared structured mapping for associating pseudo-labels with the real and fake images. Our generator outperforms unconditional GANs in terms of FID with significant margins on  large scale datasets like ImageNet and LSUN. It also outperforms class conditional GANs trained on human labels on CIFAR10 and CIFAR100 where fine-grained annotations or a large number of samples per class are not available. Additionally, our clustering network exceeds the state-of-the-art on CIFAR100 clustering.
%We achieve $8.8$ and $15.7$ on CIFAR10 and CIFAR100 respectively, compared with $11.4$ and $17.7$ for conditional GANs .

%Its features outperform their BigBiGAN counterparts by a significant margin when they are transferred for the task of object recognition.
\end{abstract}

%%%%%%%%% BODY TEXT
\section{Introduction}
Generative adversarial networks(GANs)~\cite{GANs} are known as one of the most useful subset of generative models. Standard GANs are independent of any kind of human supervision, involving adversarial training of a generator that synthesizes synthetic images given a low dimensional noise vectors, and a discriminator that aims to distinguish between synthetic and real images. They are also called unconditional GANs(uGANs), as there is no common source of information between the discriminator and the generator. Despite the high capability of GANs to generate realistic images, they are notorious for their instability and prone to mode collapse~\cite{Arjovsky17}.

Conditional GANs(cGANs)~\cite{cgan} have effectively addressed the aforementioned challenges.
In the cGAN framework, both the generator and discriminator are fed with an auxiliary input representing some factor that contributes to observations,~\eg~object labels or data from different modalities. The discriminator aims to distinguish real image/conditioning term pairs from the fake pairs. Conditional GANs guide the training by considering different modes of the data distributions. When conditioned on labels, cGANs explicitly approximate the joint distribution of observations and labels.
%consider data distribution as a mixture model where the labels determine the empirical subdistributions, \ie~each category determines a subdistribution empirically available in the dataset. 
For the sake of simplicity, we denote cGANs as those conditioned on class labels throughout the paper.  

Although cGANs benefit from a rich prior, their drawback lies in dependency on human labels.
%A well know motivation for unsupervised learning from a practical point of view is to alleviate the cost of tedious process of labeling.
To cope with this limitation, we propose Self-labeled Conditional GANs(slcGANs)
%that keep advantage of CGANs over UGANs,
%whic is being conditional,
that learn to assign labels to the images automatically by incorporating an additional clustering network. The benefits are twofold. First, training is independent of human labels. Second, the slcGAN framework provides a rich supervisory signal for our clustering network.  %This allows the learned representations by our clustering network can be further transferred to other downstream tasks, \eg~object classification.

A well known motivation for unsupervised learning from a practical point of view is to alleviate the cost and error of the labeling process. Being independent of human labels not only saves the cost of labeling but also allows us to find a presumably better set of pseudo-labels through which conditional GANs can achieve higher performance on the task of image generation. This is because human labels are based on semantic contents that still include large variety of intra-class variations.
%that might be suboptimal for the image generation task.
This issue is more problematic in datasets like CIFAR10 or LSUN~\cite{lsun}, where a fine set of labels is not available, and a large variety of images with different object and background appearances are grouped as a single class. Approximating such complex conditional distributions is still challenging. As it is shown in Figure~\ref{fig:teaser}, our clustering network splits the dataset into fine-grained clusters based on abstract factors of variation such as shape, appearance, pose, interaction with other objects and/or background, scale, quantity, etc. Each cluster proposes a simpler subdistribution that includes fewer variations and therefore is easier to approximate for the generator.  %, see Fig.~\ref{fig:1}.

Our proposed slcGAN framework includes a clustering network in addition to a standard cGAN. It assigns a $K$-dimensional probability vector to a real image as input, where $K$ is a given parameter indicating the number of clusters. An image and its corresponding class probability vector is fed to the discriminator as a real pair. The generator takes a randomly selected label,~\ie~conditioning term, in addition to a noise vector sampled from a multivariate Gaussian, and generates a fake image. A generated image and its corresponding conditioning term is fed to the discriminator as a fake pair. The clustering network and the generator aim to fool the discriminator such that it fails to distinguish between the real and fake pairs. For this purpose, the clustering network and the generator should follow the same strategy on assigning pseudo-labels to the real images and associating conditioning term to the fake images. The pseudo-label assignment strategy should facilitate the generator task of producing realistic images, as low quality images enable the discriminator to distinguish between the fake and real pairs based on the images alone regardless of the mechanism with which they are tied to the pseudo-labels.
%From a more general perspective, our method is a mixture model where the pseudo-labels are considered as latent variables. Each subdistirbution is 

A degenerate solution for the clustering network and the generator of the three player game described above is to hide any dependency between the images and the pseudo-labels. To prevent this, we explicitly enforce the generator to establish dependency between the conditioning term and the fake images. Moreover, we provide an auxiliary supervisory signal for the clustering network via multiple view clustering that enforces consistency between the pseudo-labels assigned to an image and its augmented versions. 
%We further show that the BigBiGAN frameworks~\cite{bigbigan} benefit similarly from the same supervisory signal.

Our experiments show that our proposed slcGAN framework trained with a relatively large number of clusters outperforms unconditional GANs on ImageNet and LSUN datasets with significant margins. Additionally, it outperforms cGANs trained on human labels in terms of FID on CIFAR10 and CIFAR100. 
Moreover, our clustering network exceeds the state-of-the-art on CIFAR100 in terms of clustering accuracy.
Our contributions are: I) We introduce slcGANs, a novel and fully unsupervised framework for joint clustering and  conditional GAN training. II) We show that slcGANs outperform uGANs on several large scale datasets, and cGANs conditioned on human labels in terms of FID for image generation on CIFAR10 and CIFAR100. III)  We show that our clustering network exceeds the state-of-the-art for image clustering on the CIFAR100 dataset.

%%%%%%%%%%%%%%%%%%%%%%%%%%%%%%%%%%%%%%%
%%%%%%%%%%%%%% Related work %%%%%%%%%%%
%%%%%%%%%%%%%%%%%%%%%%%%%%%%%%%%%%%%%%%
\section{Related work}
\paragraph{GANs as a source of supervisory signal.} 
%There has been recently a remarkable progress in the self-supervised learning. 
%These methods generally avoid annotations for learning representations by defining a pretext task that exploit free supervisory signals. The pretext task is defined such that the network requires to explore structure of the data to optimize the objective function. A set of successful supervisory signals includes context~\cite{contex, jigsaw}, reconstruction~\cite{Zhang:2016:colorful, Pathak:2016:context}, predicting transformations~\cite{AET, rotaion} and counting~\cite{counting}. 
%Specifically, three category of self-supervised methods are related to ours. \textit{I) Contrastive methods.} These methods have shown great performance  maximize mutual information between different views via constrastive learning. Different views are obtained via applying independent augmentations on image.~\cite{Chen:2020:SimCLR,Han:2019:DPC,Tian:2019:CMC}. Our multiple view clustering loss is related to them. Nevertheless, we are not required to prevent from degenerate solution of mapping to a fixed vector via contrastive learning. \textit{II) Clustering based methods.} Caron~\etal~\cite{caron2018deep} alternate between clustering the feature of network via k-means and predicting the pseudo-labels. Asano~\etal~\cite{asano2020self} recently explored the same direction by obtaining the pseudo-labels in each iteration via minimization cross-entropy loss using a fast variant of the Sinkhorn-Knopp algorithm.  \textitt{III) GAN-based methods.} 
GANs have been exploited before as a source of supervisory signal for learning representations. 
%The most relevant category of self-supervised methods are those that use the GAN framework for representation learning. 
Radford~\etal~\cite{DCGAN} transfer features of the discriminator of unconditional GANs. Chen~\etal~\cite{SSLGAN} investigated the same direction by training the discriminator of unconditional GANs jointly with the auxiliary task of rotation prediction~\cite{rotation}.
%Despite these methods that do not incorporate a new network other than the generator and discriminator,
Adversarially learned inference (ALI) ~\cite{ali} or bidirectional GAN(BiGAN)~\cite{bigan} proposed primarily to augment the standard GAN with an encoder network that maps real data to the latent space.
%The generator of the BiGAN framework takes a latent(noise) vector and generates a fake image, which is further fed to the discriminator with the corresponding latent vector as a fake pair. The encoder assigns a latent vector to a given real image, which is further fed to the discriminator as a real pair. The encoder and the generator aim to fool the discriminator. 
In the case of an optimal discriminator, ~\cite{bigan} showed that a deterministic BiGAN acts like an autoencoder, \ie~combination of the encoder and generator, that minimizes $\ell_0$ reconstruction costs, where  the shape of the reconstruction error surface is dictated by a parametric discriminator. Our objective is fundamentally different. We aim to decompose the distribution into smaller subdistributions, each of which is easier for the generator to approximate. The BigBiGAN framework~\cite{bigbigan} is a recently revised version of BiGAN using the BigGAN~\cite{biggan} generator and a modified discriminator.
%The BigBiGAN framework serves as the most relevant work to ours. We thoroughly compare our method to the BigBiGAN framework in terms of image generation performance as well as quality of the learned features.

\paragraph{Clustering.} End-to-end learning for image clustering has been recently explored. A line of methods~\cite{DEC, DAC, caron2018deep} follow an iterative approach that alternates between obtaining pseudo-labels via clustering, \eg~$k$-means, and feature learning by predicting the pseudo-labels obtained from the previous stage. The idea of these methods is tied to the architecture of CNNs as a prior to cluster images. Another line of research~\cite{ji2019invariant, IMSAT} exploits maximizing the mutual information between an image and its augmentations. More related to our approach, Mukherjee~\etal~\cite{clusterGAN} exploit GANs for clustering. They perform clustering in the latent space of unconditional GANs by joint training of an encoder that projects back a generated image to the corresponding latent space. Apart from the complexity of training an optimal encoder in their approach and the suboptimality of unconditional GANs, their encoder never observes real images during the training. Our clustering method obtains the supervisory signal from the generator indirectly to train a separate clustering network. Our clustering network is updated through real images, however, the training signal originates from the generated images via cooperation with the generator.

\paragraph{Conditional GANs without annotations.} Lucic~\etal~\cite{FewerLabels} exploit the feature space of a pre-trained self-supervised task along with a fraction of ground truth labels in a semi-supervised setting to alleviate the need for a fully annotated training set. Our approach requires neither a pre-trained network nor a partially labeled training set. Very recently and concurrently with us, Liu~\etal~\cite{liu2020selfconditioned} pursued a similar problem setup as ours. They perform clustering via $k$-means on the feature space of the discriminator and alternate between clustering and the conditional GAN training. This approach suffers from two main drawbacks. First, their performance relies on the quality of the intermediate representation of the discriminator, which is not necessarily an optimal feature extractor. Second, their method requires an existing clustering algorithm, \eg $k$-means. By incorporating a new clustering network to the conditional GAN framework, our method not only is independent of an existing feature extractor but also learns useful features independently of a clustering algorithm. Our experiments show that our method obtains clusters with significantly higher purity, which is essential for the image generation task.

%%%%%%%%%%%%%%%%%%%%%%%%%%%%%%%%%%%%%%%
%%%%%%%%%%%%%% Method %%%%%%%%%%%%%%%%
%%%%%%%%%%%%%%%%%%%%%%%%%%%%%%%%%%%%%%%
%%It also prevents from direct communication of $C$ and $G$...
\section{Method}
Conditioning on class labels has been effective on alleviating the mode collapse issue of GANs. They enlarge the support of the generator by guiding it to cover all the classes available in the dataset. Our proposed method aims for the same goal in an unsupervised setting in which we obtain the pseudo-labels by incorporating a clustering network.  Given a set of unlabeled images, our objective is to train simultaneously a clustering network that assigns a cluster probability vector to each datapoint used as a conditioning term for training conditional GANs. We review the conditional GAN framework in the following and discuss our method in more detail.
\subsection{Conditional GANs}
GANs involve adversarial training of two neural networks, a generator($G$) and a discriminator($D$). Given a noise vector sampled from a simple low dimensional distribution,  the generator aims to approximate the underlying unknown data distribution $\mathbb{P}_r$ to fool the discriminator, while the discriminator aims to distinguish between real and fake samples.
%In the case of an optimal discriminator, the generator we want pdata = pg, where pg is the generator distribution.
\begin{align}
\min_{G} \max_D   \quad &\,\,\mathbb{E}_{x \sim \mathbb{P}_r} [\log D(x)]  + \mathbb{E}_{z \sim p_z} [\log (1-D(G(z)))]
\label{eq:gan}
\end{align}
Conditional GANs extend the GAN framework by exploiting an auxiliary source of information available for $G$ and $D$ as an extra input,  
\begin{align}
\label{eq:cgan}
\min_{G} \max_D  \quad &\,\,\mathbb{E}_{x,y \sim \mathbb{P}_r} [\log D(x,y)] \quad  \\ \nonumber
+ \quad &  \mathbb{E}_{z \sim p_z, c\sim p_c} [\log (1-D(G(z,c),c))]
\end{align}
where the real joint samples of images and the conditioning terms are empirically available in the training dataset.  Class labels are the most widely used conditioning information in Equation~\ref{eq:cgan}. We further extend the class conditional GAN framework to not rely on explicit labels in the dataset. 

\begin{figure*}[h]
    \centering
    \includegraphics[height=.4\textwidth]{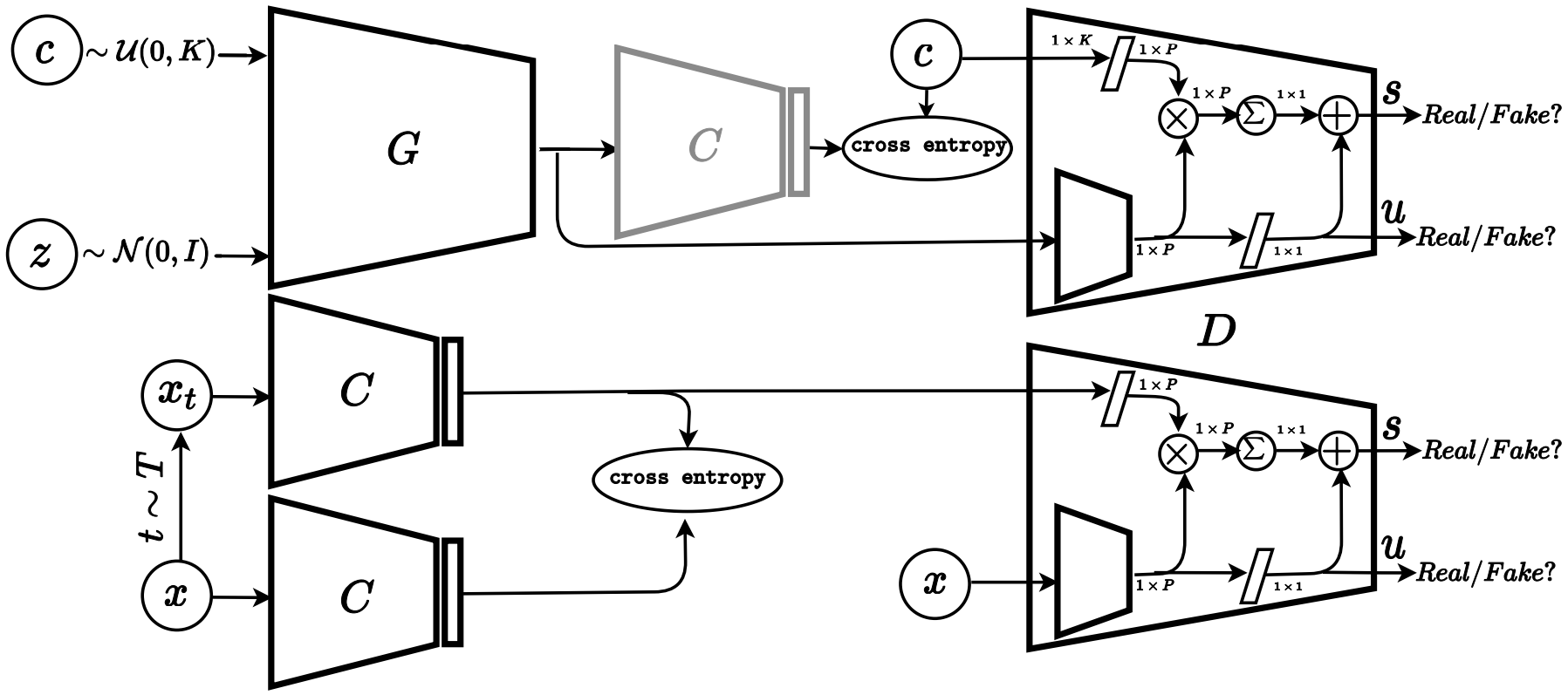}
    \caption{\footnotesize Self-labeled Conditional GANs
    \textmd{involve adversarial training between three networks. It includes a clustering network($C$) in addition to cGANs. The discriminator(D) takes an image and the corresponding pseudo-label as input. A real pair is composed of a real image and the assigned probability vector by $C$(bottom row). A fake pair is composed of a randomly generated image by the generator($G$) and the conditioning term($c$) by which it has been created(top row). $C$ and $G$ are trained against $D$ such that it fails to distinguish between real and fake pairs. Mutual information maximization is imposed via a cross entropy loss(top row). $C$ is not updated through this loss, shown in gray. Transformation consistency is imposed by a separate cross entropy loss on real samples(bottom row). Following the BigBiGAN framework, the adversarial loss involves separate terms for image($u$) and joint inputs($s$).
    }
    }
    \label{fig:method}
    \vspace{-.5cm}
\end{figure*}
%%%%%%%%%%%%%%%%%%%%%%%%%%%%%%%%%%%%%%%%%%%%%%%%%%%%%%%%%%%%%%%%%%%%%%%%%%%
%%%%%%%%%%%%%%%%%%% Self-labeled Conditional GANs %%%%%%%%%%%%%%%%%%%%%%%%%
%%%%%%%%%%%%%%%%%%%%%%%%%%%%%%%%%%%%%%%%%%%%%%%%%%%%%%%%%%%%%%%%%%%%%%%%%%%
\subsection{Self-labeled Conditional GANs}\label{sec:slcGANs}
Our proposed method involves adversarial training of three networks. In addition to $G$ and  $D$ of cGANs, we incorporate a clustering network ($C$) that assigns pseudo-labels to the images cooperating with $G$ against $D$. A primary objective function can be obtained simply by substituting labels in the cGAN framework with $p(y|x)$ approximated by $C$,
\begin{align}
\label{eq:adv_obj}
\min_{G,C} \max_D \text{ } & \mathcal{L}_{adv}(G,C,D)  = \mathbb{E}_{x \sim \mathbb{P}_r} [\log D(x,p(y|x))] \\ 
+ \quad & \mathbb{E}_{z \sim p_z, c\sim p_c} [\log (1-D(G(z,c),\mathbbm{1}_K(c)))]  \nonumber
\end{align}

where $K$ is the number of clusters, $\mathbbm{1}_K(c)$ is a $K$-dimensional one-hot encoding of $c$ sampled from prior distribution, $p_c$, and $p(y|x) =$ \texttt{softmax}$(C(x))$ is a $K$-dimensional class probability vector of $x$ determined by $C$. $p_z$ is a multivariate Gaussian and $p_c$ is a uniform distribution.
%We adopt generator and discriminator of the BigGAN framework~\cite{bigan}

To fool the discriminator, the above objective function requires $C$ and $G$ to match the joint pairs of $(x,p(y|x))$ sampled from the real distribution to the synthetic pairs of $(G(z,c),\mathbbm{1}_K(c))$. This encourages $G$ to generate realistic images, $C$ to distribute uniformly one-hot encoding probability vectors over the images, and both $C$ and $G$ to follow a shared mapping for associating pseudo-labels with the real and fake images respectively. Moreover, this mapping should assist $G$ with generating more realistic images.

\paragraph{Preventing random label assignment.} Any mismatch on pseudo-label association between $C$ and $G$ to the real and fake images provides $D$ with a signal to distinguish between the fake and real pairs. A degenerate solution for $C$ and $G$ to hide this signal from $D$ is to decouple images from the pseudo-labels. That is, random label assignment to the real images by $C$ and independent fake images of the conditioning term by $G$.
%Such a degenerate solution turns our method into the uGAN framework.

To prevent this degenerate solution, we explicitly enforce the fake images to depend on the conditioning term. We follow the same approach as the InfoGAN framework~\cite{infogan} by maximizing a variational lower bound of $I(c;G(z,c))$  parameterized by a neural network, where $I$ denotes mutual information. The InfoGAN framework adapts the discriminator for this purpose by adding a classification head for the fake images in addition to the one for real/fake discrimination.
%\footnote{Any classifier works similarly here. InfoGAN uses $D$ to avoid any additional computational cost.}.

We however can not follow the same approach as our discriminator is conditional. It takes $c$ as the second input, which makes predicting $c$ a trivial task. We thus use $C$ alternatively by adding the following loss function:
\begin{align}
\min_{G} & \text{  } \mathcal{L}_{mi}(G)  = \mathbb{E}_{z \sim p_z, c\sim p_c} -\log[p(y=c|G(z,c))]
\label{eq:mi}
\end{align}
where $p(y=c|G(z,c))$ is estimated by $C$ as it is performed for the real images in Equation~\ref{eq:adv_obj}. Note that $C$ can also be updated by $\mathcal{L}_{mi}$. However, we obtained better performance by training this loss only for $G$. One reason could be that direct communication between $G$ and $C$ yields degenerate solutions by encoding shortcuts in the fake images by $G$. These shortcuts allow $G$ and $C$ to minimize $\mathcal{L}_{mi}$  without establishing a comprehensive dependency between the fake images and the conditioning term. Furthermore,  optimizing $\mathcal{L}_{mi}$ only for $G$ provides the generator with another supervisory signal for generating more realistic images consistent with $C$. Intuitively, it encourages $G$ to generate realistic images such that $C$ behaves on them in a similar manner to real images in terms of pseudo-label prediction. Note that $C$ is only updated via real images throughout the training. 
%That is, the fake images should be classified correctly by a classifier that is only updated through real inputs. 

\paragraph{Multiple view clustering.}
%The above objective function works reasonably well on simple distributions,~\eg MNIST. For  more complex distributions that suffer from less accurate gradients, 
We further provide an auxiliary supervisory signal for $C$ via multiple view clustering. That is, $C$ should assign the same pseudo-label to an image and its augmented versions,
\begin{align}
\min_{C}  \mathcal{L}_{aug}(C) =  \mathbb{E}_{x \sim \mathbb{P}_r} \sum_{c=1}^{K} -p(y=c|x_t) \log[p(y=c|x)].
\label{eq:aug}
\end{align}
Where $x_t$ is a randomly augmented version of $x$ via cropping, color jittering, and flipping. 

Our approach of multiple view clustering is inline with learning via invariances across augmentations~\cite{Chen:2020:SimCLR}, an approach that has shown significant improvements in the field of self-supervised representation learning. These methods mostly rely on contrastive learning~\cite{infonce} to prevent the degenerate solution of constant prediction, which is not required in our method. The supervisory signal originating from the adversarial training enforces $C$ to mimic a similar distribution as $p_c$, which is simply a uniform distribution. Our final objective function is the following. 
\begin{align}
\min_{G,C} \max_D  \quad  \lambda_1 \mathcal{L}_{adv}(G,C,D) + \lambda_2 \mathcal{L}_{mi}(G) + \lambda_3 \mathcal{L}_{aug}(C)
\label{eq:all}
\end{align}
where $\lambda_i$ denotes the corresponding coefficient for each loss, and are set to $1$ during training. Figure~\ref{fig:method} illustrates the slcGAN framework components and its three loss functions.

\begin{algorithm}[H]
	\footnotesize
	\caption{\footnotesize Training steps of the slcGAN training. The adversarial loss for updating $D$ is split across unary and joint terms(line $14,15$).$G$ is updated via the adversarial loss in line $20$ and mutual information maximization loss in line $21$. $C$ is updated for the adversarial and multiple view clustering loss in lines $27,28$ respectively.}\label{euclid}
	\begin{algorithmic}[1]
		\Procedure {SAMPLEDATA}{}
		\State $B_x = \{x^{(1)}, \dots , x^{(n)}\} \leftarrow $ A batch of real images of size $n$
		\State $B_{z} =  \{z^{(1)}, \dots , z^{(n)}\} \leftarrow$ A batch of random samples from a
		\State $\qquad \qquad  \qquad \qquad \qquad \quad$ Gaussian of size $n$
		\State $B_{c} =  \{c^{(1)}, \dots , c^{(n)}\} \leftarrow$ A batch of random one hot encoded
		\State $\qquad \qquad  \qquad \qquad \qquad \quad$ labels of size $n$
\State \Return $B_x, B_z, B_c$
\EndProcedure

\For{number of training iterations}
\State $B_x, B_z, B_c =  SAMPLEDATA()$
\State Update discriminator via:\\
$\qquad \quad u^{(i)}_r, s^{(i)}_r = D(x^{(i)}, C(x^{(i)}))$ \\
$\qquad \quad u^{(i)}_f, s^{(i)}_f = D(G(z^{(i)},c^{(i)}), c^{(i)})$ \\
$\qquad \quad \nabla_{\theta_d} \frac{1}{n}\sum_{i=1}^{n} [\max(0,1-u^{(i)}_r) + \max(0,1-s^{(i)}_r)$ \\ \nonumber $\qquad \qquad \qquad \qquad + \max(0,1+u^{(i)}_f) + \max(0,1+s^{(i)}_f)]$
\State $\sim, B_z, B_{c} =  SAMPLEDATA()$
\State Update generator via:\\
$\qquad \quad u^{(i)}_f, s^{(i)}_f = D(G(z^{(i)},c^{(i)}), c^{(i)})$ \\
$\qquad \quad y^{(i)} =  \texttt{softmax}\{C(G(z^{(i)},c^{(i)}))\}$ \\
$\qquad \quad \nabla_{\theta_g} \frac{1}{n}\sum_{i=1}^{n} [ -u^{(i)}_f - s^{(i)}_f$]\\
$\qquad \quad \nabla_{\theta_g} \frac{1}{n}\sum_{i=1}^{n} [\sum_{j=1}^{k}  -c^{(i)}_{j} \log y^{(i)}_{j}]$
\State $B_x, \sim, \sim  =  SAMPLEDATA()$
\State Update clustering via:\\
$\qquad \quad \sim, s^{(i)}_r = D(x^{(i)}, C(x^{(i)}))$ \\
$\qquad \quad x^{(i)}_t =  T(x^{(i)})$ \\
$\qquad \quad p^{(i)} =  \texttt{softmax}\{C(x^{(i)})\},  q^{(i)} =  \texttt{softmax} \{C(x^{(i)}_t)\}$   \\
$\qquad \quad \nabla_{\theta_c} \frac{1}{n}\sum_{i=1}^{n} [s^{(i)}_r]$\\
$\qquad \quad \nabla_{\theta_c} \frac{1}{n}\sum_{i=1}^{n} [\sum_{j=1}^{k}  -q^{(i)}_{j} \log p^{(i)}_{j}]$
\EndFor
\end{algorithmic}
\label{algo:training}
%\vspace{-.1cm}
\end{algorithm}

\paragraph{Architecture and adversarial losses.}
Our generator and discriminator architecture is based on those of BigGAN~\cite{biggan}. 
%In a supervised setting, class labels are provided via positive integers. BigGAN maps the class labels to vectors using word embeddings . These operations are not applicable in our discriminator since a non-differentiable \texttt{argmax} is required.
We use a linear transformation to map pseudo-class probability vectors to higher dimensional dense vectors that are later provided to $G$ with class-conditional BatchNorm~\cite{Dumoulin17, Vries17}, and to $D$ with projection~\cite{MiyatoProjection}.  We use spectral normalization ~\cite{MiyatoSpectral} in both $C$ and $G$ and train the adversarial objective functions using the hinge loss~\cite{Hyun17,Tran17}. Following the BigBiGAN framework, we include a unary term in our adversarial loss that is a function of only images, and split the loss values across two terms. The unary term encourages $G$ towards explicitly generating realistic images, which is in turn implicitly performed via the joint term,~\ie~Equation~\ref{eq:adv_obj}. Algorithm~\ref{algo:training} demonstrates the training steps of slcGANs in detail.

%%%%%%%%%%%%%%%%%%%%%%%%%%%%%%%%%%%%%%%%%%%%%%%%%%%%%%%%%%%%%%%%%%
%%%% Experiments 
%%%%%%%%%%%%%%%%%%%%%%%%%%%%%%%%%%%%%%%%%%%%%%%%%%%%%%%%%%%%%%%%
\section{Experiments}
We evaluate our generator in terms of image generation performance and the accuracy of our clustering network. Ground truth labels provide a plausible upper bound for performance in both cases. We conduct experiments on MNIST, CIFAR10, CIFAR100, LSUN, and ImageNet datasets.

\textbf{Implementation details.} 
We set the channel multiplier to $128$ and the number of residual blocks per stage to $1$ for both $G,D$. We use standard ResNet18~\cite{resnet} as the backbone of our clustering network, Adam~\cite{Adam} optimizer with a batch size of $256$ for all networks, a learning rate of $1e-4$, and update the discriminator twice per generator update. 

\subsection{Image synthesis}
We evaluate image generation performance quantitatively by reporting the Inception Score(IS)~\cite{insecption} and the Fréchet Inception Distance(FID)~\cite{FID}.
The unconditional GAN, class conditional GAN, and slcGAN are trained with the same architecture. We evaluate our method for small, medium, and large values of $K$, and perform qualitative evaluations later in section~\ref{subsec:qual}.
\paragraph{CIFAR.}  We train our model with $(10,50,100)$ on CIFAR10 and $(20,100,500)$ on CIFAR100 as small, medium, and large values of $K$ respectively. As it is shown in Table~\ref{tbl:cifar}, the slcGAN framework outperforms the unsupervised baselines as well as the cGAN by significant margins. We conjecture that this is due to the number of modes imposed by a large number of clusters, yielding more variant synthetic images, and the flexibility of associating pseudo-labels to the fake images by the generator compared to preassigned labels.
\begin{table}[h]
    %\footnotesize
    \begin{adjustbox}{width=\linewidth}
    \centering
    \begin{tabular}{l  c c c  c c  }
    \toprule
     \multirow{2}{*}{} &  \multicolumn{2}{c}{\textbf{CIFAR10}} & & \multicolumn{2}{c}{\textbf{CIFAR100}}
     \\\cline{2-3} \cline{5-6}
     & IS($\uparrow$) & FID($\downarrow$) & & IS($\uparrow$) & FID($\downarrow$) \\
     \midrule
     uGAN        &  $7.65 \pm 0.03$ & $11.42 \pm 0.19$ & & $7.54\pm 0.04 $ & $18.31 \pm 0.48$\\
     Liu~\etal~\cite{liu2020selfconditioned}   &  $7.72 \pm 0.03$ & $18.03 \pm 0.55$ & & - &-\\
     slcGAN(small $K$) &  $7.46 \pm 0.09$ & $13.09 \pm 0.97$ & & $7.48 \pm 0.06$ & $\textbf{13.96} \pm 0.38$ \\
     slcGAN(medium $K$) &  $\textbf{8.07} \pm 0.07$ & $\textbf{8.95} \pm 0.55$ & & $\textbf{7.54} \pm 0.07$ & $15.36 \pm 0.42$ \\
     slcGAN(large $K$)&   $7.98 \pm 0.09$ & $12.2 \pm 0.38$ & & $6.67 \pm 0.06$ & $23.95 \pm 1.67$  \\
     \midrule
     cGAN        &  $9.12 \pm 0.06 $ & $11.31 \pm 1.51$   & & $8.46 \pm 0.07$ & $17.28 \pm 2.25$ \\
     %slcSLCGAN$_{100}$ - & - & & 9.12 & 11.31 & & - & - \\
     \bottomrule
    \end{tabular}
    \end{adjustbox}
    \caption{\footnotesize Image generation performance on CIFAR10 and CIFAR100.
    \textmd{ slcGAN is trained with  $(10,50,100)$ on CIFAR10 and $(20,100,500)$ on CIFAR100 respectively. 
    %The best number per row is bold faced. 
    We outperform cGANs trained on ground truth labels in terms of FID by significant margins on both datasets. Results are averaged over five trials.
    }
    }
    \label{tbl:cifar}
\end{table}

\paragraph{LSUN and ImageNet.}
We train our model on $64\times 64$ images of the ILSVRC~\cite{imagenet_cvpr09} training set and $2M$ randomly selected images of $20$ object categories of LSUN~\cite{lsun}, $100K$ per category. We use   $(20,100,1000)$ for LSUN and $(100,1000,2000)$ for ImageNet as small, medium, and large values of $K$ respectively. For evaluation, we use the validation set of ILSVRC and $200K$ randomly selected images of $20$ object categories of LSUN~\cite{lsun}, $10K$ per category, and separate from the training set. Table~\ref{tbl:lsun_imagenet} summarizes the results. Our method outperforms the unsupervised baseline, closing the gap with the ground truth labels. Our generator achieves FID of $14.22, 19.24$ on LSUN and ImageNet respectively, which are significant improvements compared to $25.68, 28.69$ of the unconditional GAN. 

\begin{table}[h]
    %\footnotesize
    \begin{adjustbox}{width=\linewidth}
    \centering
    \begin{tabular}{l  c c c     c c  }
    \toprule
     \multirow{2}{*}{} &  \multicolumn{2}{c}{\textbf{LSUN}} & & \multicolumn{2}{c}{\textbf{ImageNet}}
     \\\cline{2-3} \cline{5-6}
     & IS($\uparrow$) & FID($\downarrow$) & & IS($\uparrow$) & FID($\downarrow$) \\
     \midrule
     uGAN        &  $9.60$ & $25.68$ & & $11.71$ & $28.69$\\
     slcGAN(small $K$) &  $10.28$ & $15.49$ & & $14.43$ & $19.71$ \\
     slcGAN(medium $K$) &  $10.66$ & $14.71$ & & $\textbf{16.48}$ & $\textbf{19.24}$ \\
     slcGAN(large $K$)&   $\textbf{11.05}$ & $\textbf{14.22}$ & & $14.37$ & $22.47$  \\
     \midrule
     cGAN        &  $11.36$ & $8.47$   & & $25.60$ & $10.84$ \\
     %slcSLCGAN$_{100}$ - & - & & 9.12 & 11.31 & & - & - \\
     \bottomrule
    \end{tabular}
    \end{adjustbox}
    \caption{\footnotesize Image generation performance on LSUN and ImageNet.
    \textmd{ slcGAN is trained with $(20,100,1000)$ for LSUN and $(100,1000,2000)$ as small, medium, and large values of $K$ respectively. 
    %The best number per row is bold faced. 
    We outperform the unsupervised baselines, closing the gap with the ground truth labels.
    }
    }
    \label{tbl:lsun_imagenet}
\end{table}

%%%%%%%%%%%%%%%%%%%%%%%%%%%%%%%%%%%%%% Clustering %%%%%%%%%%%%%%%%%%%%%%%%%%%%%
\subsection{Clustering}
We evaluate the clustering accuracy when the number of clusters matches the ground truth. For CIFAR100, we consider two cases of $20$ superclasses and the original fine-grained $100$ classes.
Although slcGAN does not outperform the baselines on MNIST and CIFAR10, and $20$ superclasses on CIFAR100, it outperforms  IIC~\cite{ji2019invariant} on CIFAR100 when the ground truth includes fine-grained labeling, $15.7\%$ versus $9.5\%$.\footnote{IIC~\cite{ji2019invariant} does not report CIFAR100 clustering with $100$ clusters. The authors provided us with this result.} 
We believe this is related to the better generator performance on approximating smaller subdistributions represented by each fine-grained cluster. Our method benefits from a large number of clusters as our generator provides the clustering network with more accurate gradients via the discriminator.
The results in Table~\ref{tbl:clus} shows the superiority of our method on fine-grained clustering.
\begin{table}[h]
    \begin{adjustbox}{width=\linewidth}
    \begin{tabular}{l c  c  c c }
    \toprule
   & \multirow{2}{*}{\textbf{MNIST}} & \multirow{2}{*}{\textbf{CIFAR10}} &   \multicolumn{2}{c}{\textbf{CIFAR100}}  \\\cline{4-5}
   & & & $k=20$ & $k=100$\\
     \midrule
    DCGAN~\cite{DCGAN} & $82.8$ & $31.5$ & $15.1$ & - \\
    DAC~\cite{DAC}  & $97.8$ & $52.2$ & $23.8$ & - \\
    ADC  ~\cite{ADC} & $99.2$ & $32.5$ & $16.0$ & - \\
    DeepCluster ~\cite{caron2018deep} & $65.6$ & $37.4$ & $18.9$  & - \\
    IIC ~\cite{ji2019invariant} & $\textbf{99.2}$ & $\textbf{61.7}$ & $\textbf{25.7}$ & $9.5$ \\
    %BigBiGAN~\cite{bigbigan}  & ? & ? & ? & ?  \\
    %BigBiGAN$_{mv}$  & ? & ? & ? & ?  \\
    slcGAN & $97.3$ & $47.1$ & $21.5$ & $\textbf{15.7}$\\
    %slcSLCGAN$_{100}$ - & - & & 9.12 & 11.31 & & - & - \\
    \bottomrule
    \end{tabular}
    \end{adjustbox}
    \caption{\footnotesize Clustering evaluations.
    \textmd{We find the best permutation that maps our pseudo-labels to the ground truth and measure the accuracy. The numbers of other methods are taken from~\cite{ji2019invariant}. We significantly outperform IIC on fine-grained clustering on CIFAR100. 
    }
    }
    \label{tbl:clus}
    \hspace{0.02\linewidth}
    \vspace{-.7cm}
\end{table}

\paragraph{Overclustering.} 
Our generator favors a fine-grained clustering where each cluster represents specific attributes of data. A plausible assessment for this purpose is purity in terms of ground truth labels as images with the same ground truth label are more likely to share the same fine-grained attributes. 
Purity involves assigning each cluster to the class which is most frequent in the cluster and then measuring the accuracy of this assignment by counting the number of correctly assigned datapoints. It is computed as $\frac{1}{N} \sum_k \max_j |\pi_k \cap c_j|$, where  $\Pi = \{\pi_1, \dots, \pi_K\}$ is the set of clusters, $C = \{c_1, \dots, c_J\}$ is the ground truth classes, and $N$ is the dataset size. Given a dataset with a coarse set of classes, a higher purity indicates dividing each class into fine pseudo-classes. 

Purity is affected by the $K$ value as well as cluster distribution prior, \ie $p_c$ in Equation~\ref{eq:adv_obj}, which both are unknowns. To obtain a better insight into how semantics is involved in the clustering, we train a linear classifier on the $512$ dimensional features of the layer before the last for image classification using the ground truth labels. The higher performance indicates more semantics involved in the intermediate layers of $C$ for clustering images. As it is shown in Table~\ref{tbl:purity}, purity correlates with linear separability, and our method achieves an acceptable performance, Liu~\etal~\cite{liu2020selfconditioned} achieves a purity of $11.73$ on CIFAR10 with $K=100$.

\begin{table}[h]
    %\footnotesize
    \begin{adjustbox}{width=\linewidth}
    \centering
    \begin{tabular}{l  c c c     c c  }
    \toprule
     \multirow{2}{*}{} &  \multicolumn{2}{c}{\textbf{Purity}} & & \multicolumn{2}{c}{\textbf{Linear Separability}}
     \\\cline{2-3} \cline{5-6}
     & CIFAR10 & LSUN & & CIFAR10 & LSUN \\
     \midrule
     Random initialization  &  $-$ & $-$ & & $31.8$ & $20.4$ \\
     Liu~\etal~\cite{liu2020selfconditioned}  &  $11.7$ & $-$ & & $-$ & $-$ \\
     slcGAN(small $K$)  &  $47.1$ & $24.7$ & & $61.0$ & $48.8$ \\
     slcGAN(medium $K$) &  $\textbf{52.5}$ & $41.3$ & & $\textbf{68.3}$ & $\textbf{60.5}$ \\
     slcGAN(large $K$)  &  $51.4$ & $\textbf{46.7}$ & & $65.2$ & $58.7$  \\
     \midrule
     supervised        &  - & -   & & 92.3 & 79.1 \\
     %slcSLCGAN$_{100}$ - & - & & 9.12 & 11.31 & & - & - \\
     \bottomrule
    \end{tabular}
    \end{adjustbox}
    \caption{\footnotesize Purity and Linear separability analyses.
    \textmd{Our method is trained with $(10,50,100)$ for CIFAR10 and $(20,100,1000)$ for LSUN as small, medium, and large values of $K$ respectively. The linear classifiers are trained using cross-entropy with softmax activation on the $512$ dimensional features of the layer before the last. We use no non-linearity such as batch normalization during training nor test time.
    }
    }
    \label{tbl:purity}
\end{table}

\begin{figure*}[h]
\centering
\includegraphics[width=\textwidth]{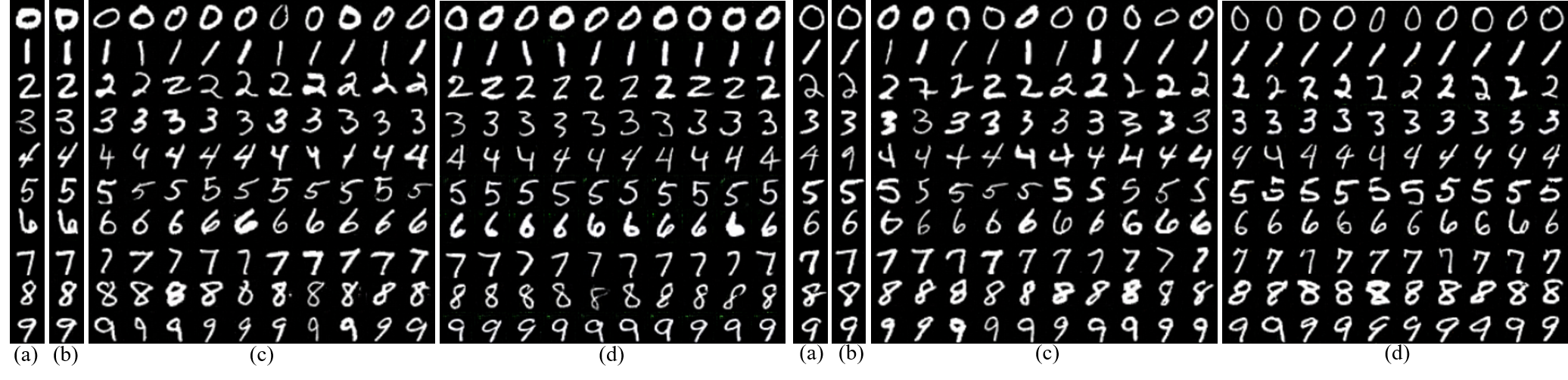}
\caption{\footnotesize Qualitative comparison with BigBiGAN on MNIST.\textmd{ (a) shows the input image, (b) shows the BigBiGAN reconstruction, (c) and (d) show 10 randomly generated samples by our model trained with $10$ and   $50$ pseudo-classes respectively.}}
\label{fig:rec}
\vspace{-.2cm}
\end{figure*}

\begin{figure}[h]
\includegraphics[trim = 12 0 35 28, clip,width=\linewidth]{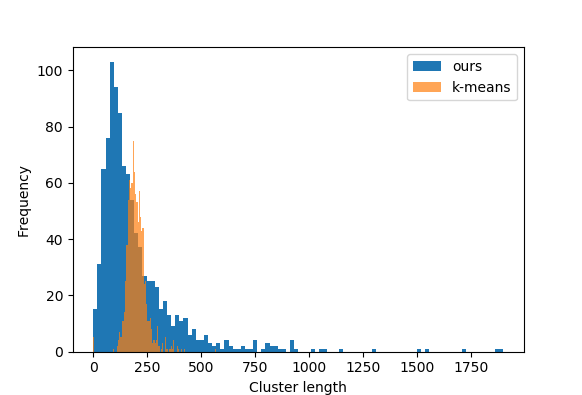}
\vspace{-.5cm}
\caption{\footnotesize{We study the impact of uniform distribution prior by comparing the histogram of clusters length of our method with $k$-means applied on the feature of the layer before the last. Our method covers a broader range of cluster lengths than $k$-means.}}
\label{fig:clustering_hist}
\end{figure}

\paragraph{Clusters distribution.} As it is mentioned in section~\ref{sec:slcGANs}, our objective function in Equation~\ref{eq:adv_obj} encourages $C$ to distribute uniformly one-hot encoding probability vectors over the images. Note that we do not benefit from true ground truth distribution here, since the number of clusters in our training is relatively higher than ground truth. Moreover, it turns out that imposed constraint via adversarial training is not strong. Figure~\ref{fig:clustering_hist} compares our method with $k$-means in terms of the histogram of the clusters length, the number of samples assigned to the clusters. Our method is trained on LSUN with $K=1000$, and $k$-means is applied on the features of the layer before the last for the same number of clusters. Our method covers a broader range of clusters length than $k$-means. This observation indicates that our method benefits from the weekly imposed constraint via adversarial training to adjust the distribution of the clusters to facilitate the image generation task of the generator.

%%%%%%%%%%%%%%%%%%%%%%%%%%%%%%%%%%%%%% Ablation studies %%%%%%%%%%%%%%%%%%%%%%%%
\paragraph{Impact of the multiple view clustering.} Our clustering network exploits two supervisory signals. The main one originates from the generator and is back propagated via the discriminator. The auxiliary signal is provided via the multiple view clustering loss, \ie~Equation~\ref{eq:aug}. To evaluate the impact of this auxiliary signal, we train our model on CIFAR10 without the multiple view clustering loss using $50$ clusters.  Our clustering features achieve $63.7\%$ top-1 accuracy via liner classification on top of the res4 layer. A drop of $5\%$, compared to $68.3\%$ shown in Table~\ref{tbl:purity}, indicates that the multiple view clustering loss is effective, and more importantly that the main signal originates from the adversarial training. 

\subsection{Qualitative evaluation}\label{subsec:qual}
\begin{figure*}[htb]
		    \begin{minipage}{.499\textwidth}
			\begin{minipage}[t]{.496\textwidth}
				\centering
				Generated cluster
			\end{minipage}
			\begin{minipage}[t]{.496\textwidth}
				\centering
				Real cluster
			\end{minipage}
			
			\vspace{.1cm}
			\begin{minipage}{.496\textwidth}
				\includegraphics[trim=0 0 0 0, clip,width=1\textwidth]{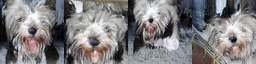}
				
				\includegraphics[trim=0 0 0 0, clip,width=1\textwidth]{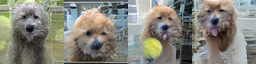}
				
				\includegraphics[trim=0 0 0 0, clip,width=1\textwidth]{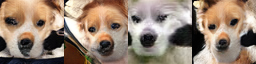}
				
				\includegraphics[trim=0 0 0 0, clip,width=1\textwidth]{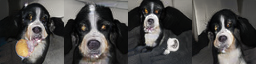}

				\includegraphics[trim=0 0 0 0, clip,width=1\textwidth]{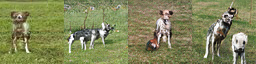}
				
				\includegraphics[trim=0 0 0 0, clip,width=1\textwidth]{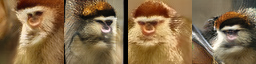}
				
				\includegraphics[trim=0 0 0 0, clip,width=1\textwidth]{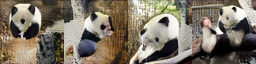}
				
				\includegraphics[trim=0 0 0 0, clip,width=1\textwidth]{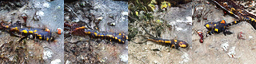}
			\end{minipage}
			\begin{minipage}{.496\textwidth}
			    \includegraphics[trim=0 0 0 0, clip,width=1\textwidth]{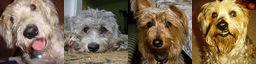}
				
				\includegraphics[trim=0 0 0 0, clip,width=1\textwidth]{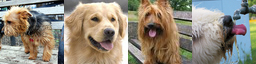}
				
				\includegraphics[trim=0 0 0 0, clip,width=1\textwidth]{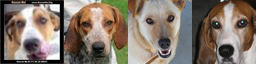}
				
				\includegraphics[trim=0 0 0 0, clip,width=1\textwidth]{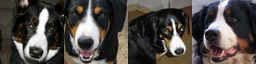}

				\includegraphics[trim=0 0 0 0, clip,width=1\textwidth]{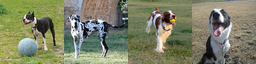}
				
				\includegraphics[trim=0 0 0 0, clip,width=1\textwidth]{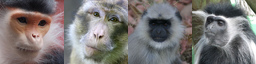}
				
				\includegraphics[trim=0 0 0 0, clip,width=1\textwidth]{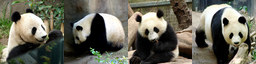}
				
				\includegraphics[trim=0 0 0 0, clip,width=1\textwidth]{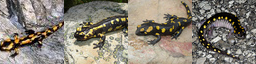}

			\end{minipage}
	    \end{minipage}
	  %}
	  %\setlength{\fboxsep}{1pt}\framebox{
		\begin{minipage}{.499\textwidth}
			\begin{minipage}[t]{.496\textwidth}
				\centering
				Generated cluster
			\end{minipage}
			\begin{minipage}[t]{.496\textwidth}
				\centering
				Real cluster
			\end{minipage}
			
			\vspace{.1cm}
			\begin{minipage}{.496\textwidth}
			    \includegraphics[trim=0 0 0 0, clip,width=1\textwidth]{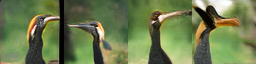}
			    
			    \includegraphics[trim=0 0 0 0, clip,width=1\textwidth]{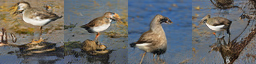}
			    
				\includegraphics[trim=0 0 0 0, clip,width=1\textwidth]{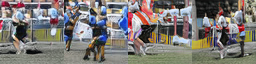}
				
				\includegraphics[trim=0 0 0 0, clip,width=1\textwidth]{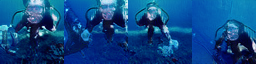}
				
				\includegraphics[trim=0 0 0 0, clip,width=1\textwidth]{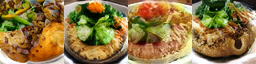}
				
				\includegraphics[trim=0 0 0 0, clip,width=1\textwidth]{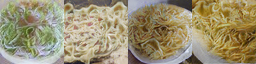}
				
				\includegraphics[trim=0 0 0 0, clip,width=1\textwidth]{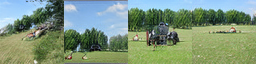}
				
				\includegraphics[trim=0 0 0 0, clip,width=1\textwidth]{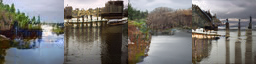}
			\end{minipage}
			\begin{minipage}{.496\textwidth}
			    \includegraphics[trim=0 0 0 0, clip,width=1\textwidth]{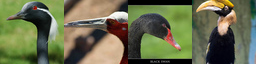}
			
			    \includegraphics[trim=0 0 0 0, clip,width=1\textwidth]{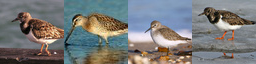}
			
				\includegraphics[trim=0 0 0 0, clip,width=1\textwidth]{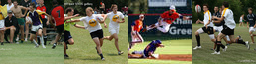}
				
				\includegraphics[trim=0 0 0 0, clip,width=1\textwidth]{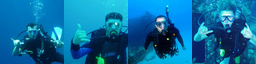}
				
				\includegraphics[trim=0 0 0 0, clip,width=1\textwidth]{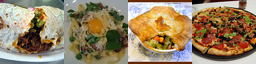}
				
				\includegraphics[trim=0 0 0 0, clip,width=1\textwidth]{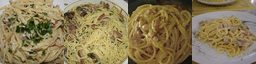}
			
				\includegraphics[trim=0 0 0 0, clip,width=1\textwidth]{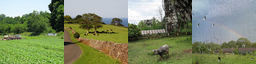}
				
				\includegraphics[trim=0 0 0 0, clip,width=1\textwidth]{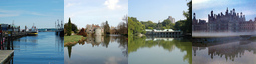}

			\end{minipage}
	    \end{minipage}
	%}
\caption{Visualization of the generated and real images of several clusters for ImageNet. Each row shows a cluster and the generated images conditioned on the same pseudo-label.}
\label{fig:qual_imagenet}
\end{figure*}

\begin{figure*}[htb]
		    \begin{minipage}{.499\textwidth}
			\begin{minipage}[t]{.496\textwidth}
				\centering
				Generated cluster
			\end{minipage}
			\begin{minipage}[t]{.496\textwidth}
				\centering
				Real cluster
			\end{minipage}
			
			\vspace{.1cm}
			\begin{minipage}{.496\textwidth}
				\includegraphics[trim=0 0 0 0, clip,width=1\textwidth]{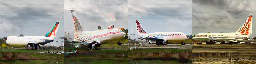}
				
				\includegraphics[trim=0 0 0 0, clip,width=1\textwidth]{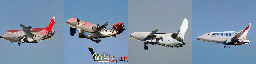}
				
				\includegraphics[trim=0 0 0 0, clip,width=1\textwidth]{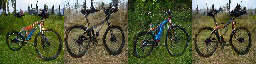}
				
				\includegraphics[trim=0 0 0 0, clip,width=1\textwidth]{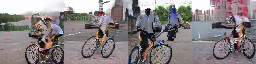}
				
				\includegraphics[trim=0 0 0 0, clip,width=1\textwidth]{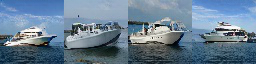}
				
				\includegraphics[trim=0 0 0 0, clip,width=1\textwidth]{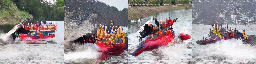}
				
			\end{minipage}
			\begin{minipage}{.496\textwidth}
				\includegraphics[trim=0 0 0 0, clip,width=1\textwidth]{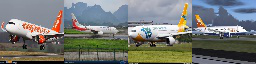}
				
				\includegraphics[trim=0 0 0 0, clip,width=1\textwidth]{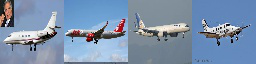}
				
				\includegraphics[trim=0 0 0 0, clip,width=1\textwidth]{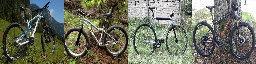}
				
				\includegraphics[trim=0 0 0 0, clip,width=1\textwidth]{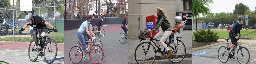}
				
                \includegraphics[trim=0 0 0 0, clip,width=1\textwidth]{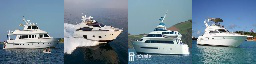}
                
                \includegraphics[trim=0 0 0 0, clip,width=1\textwidth]{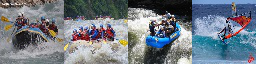}
			\end{minipage}
	    \end{minipage}
	  %}
	  %\setlength{\fboxsep}{1pt}\framebox{
		\begin{minipage}{.499\textwidth}
			\begin{minipage}[t]{.496\textwidth}
				\centering
				Generated cluster
			\end{minipage}
			\begin{minipage}[t]{.496\textwidth}
				\centering
				Real cluster
			\end{minipage}
			
			\vspace{.1cm}
			\begin{minipage}{.496\textwidth}

				\includegraphics[trim=0 0 0 0, clip,width=1\textwidth]{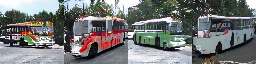}
				
				\includegraphics[trim=0 0 0 0, clip,width=1\textwidth]{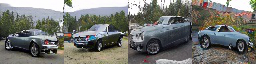}
				
				\includegraphics[trim=0 0 0 0, clip,width=1\textwidth]{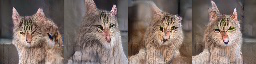}
				
				\includegraphics[trim=0 0 0 0, clip,width=1\textwidth]{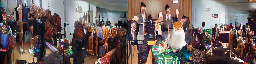}
				
				\includegraphics[trim=0 0 0 0, clip,width=1\textwidth]{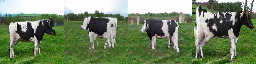}
				
				\includegraphics[trim=0 0 0 0, clip,width=1\textwidth]{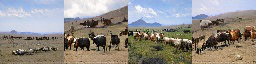}
			\end{minipage}
			\begin{minipage}{.496\textwidth}
				
				\includegraphics[trim=0 0 0 0, clip,width=1\textwidth]{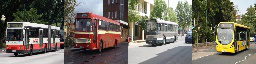}
				
				\includegraphics[trim=0 0 0 0, clip,width=1\textwidth]{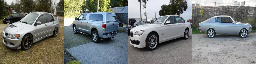}
				
				\includegraphics[trim=0 0 0 0, clip,width=1\textwidth]{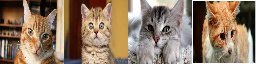}
				
				\includegraphics[trim=0 0 0 0, clip,width=1\textwidth]{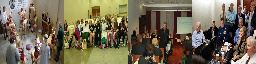}
				
				\includegraphics[trim=0 0 0 0, clip,width=1\textwidth]{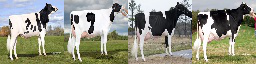}
				
				\includegraphics[trim=0 0 0 0, clip,width=1\textwidth]{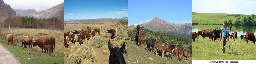}
			\end{minipage}
	    \end{minipage}
	%}
\caption{Visualization of the generated and real images of several clusters for LSUN 20 object categories. Each row shows a cluster and the generated images conditioned on the same pseudo-label.}
\label{fig:qual_lsun}
\end{figure*}

%We show images with highest and lowest confidence for some clusters in Figure~\ref{fig:cluster_sort_conf} to support our discussion on cluster purification above.
We show several synthesised and real clusters  of ImageNet in Figure~\ref{fig:qual_imagenet} and LSUN in Figure~\ref{fig:qual_lsun}. Our model is trained on both datasets with $1000$ clusters. More qualitative results are available in the supplementary material.

\paragraph{Reconstruction comparison with BigBiGAN.} Despite that the BigBiGAN framework can act like an autoencoder, \ie~ a combination of the encoder and generator, our method does not reconstruct the copy of a given image. However, our model allows us to reconstruct multiple instances of the same cluster that a given image belongs to. That is, we obtain the pseudo-label of a given image via the clustering network and feed it to the generator with multiple randomly sampled noise vectors. In the case of fine-grained labels, we are able to reconstruct multiple instances with the same style of a given image. Figure~\ref{fig:rec} shows some examples on the MNIST dataset. Given an image (a), we show BigBiGAN reconstruction (b), samples generated by our model trained with $10$ pseudo-classes  (c), and $50$ pseudo-classes (d). For each digit we show two samples with different styles. The results show that a fine-grained model recovers the same style of the input, while the coarse-grained model recovers images with the same category. BigBiGAN reconstructs the exact input image.

%\end{comment}
\section{Conclusion}
We have shown that class conditional GANs can be trained independently of the human labels, achieving higher performance than unconditional GANs. Our main idea involves incorporating a clustering network into the class conditional GAN framework that automatically assigns pseudo-labels to the images. The benefits of our proposed method are twofold. Our framework not only improves image generation performance on unlabeled datasets but also achieves a useful clustering network that exceeds the state-of-the-art for fine-grained clustering. We believe our work develops further the active line of research of generative models towards higher quality and annotation free frameworks. Moreover, we show GANs provide rich supervisory signals for clustering and representation learning.

\paragraph{Acknowledgement.} I thank Dan Zhang and Chaithanya Kumar Mummadi for useful discussions, William Beluch and Nadine Behrmann for their feedbacks on the text, and Xu Ji for providing their clustering result on CIFAR100.

{\small
\bibliographystyle{ieee_fullname}
\bibliography{slcgan}

\begin{thebibliography}{10}\itemsep=-1pt

\bibitem{Arjovsky17}
Martin Arjovsky and Léon Bottou.
\newblock Towards principled methods for training generative adversarial
  networks.
\newblock In {\em ICLR}, 2017.

\bibitem{bigan}
Andrew Brock, Jeff Donahue, and Karen Simonyan.
\newblock Large scale gan training for high fidelity natural image synthesi.
\newblock In {\em ICLR}, 2019.

\bibitem{caron2018deep}
Mathilde Caron, Piotr Bojanowski, Armand Joulin, and Matthijs Douze.
\newblock Deep clustering for unsupervised learning of visual features.
\newblock In {\em ECCV}, 2018.

\bibitem{DAC}
Jianlong Chang, Lingfeng Wang, Gaofeng Meng, Shiming Xiang, and Chunhong Pan.
\newblock Deep adaptive image clustering.
\newblock In {\em ICCV}, 2017.

\bibitem{Chen:2020:SimCLR}
Ting Chen, Simon Kornblith, Mohammad Norouzi, and Geoffrey Hinton.
\newblock A simple framework for contrastive learning of visual
  representations.
\newblock In {\em arXiv}, 2020.

\bibitem{SSLGAN}
Ting Chen, Xiaohua Zhai, Marvin Ritter, Mario Lucic, and Neil Houlsby.
\newblock Self-supervised gans via auxiliary rotation loss.
\newblock In {\em CVPR}, 2019.

\bibitem{infogan}
Xi Chen, Yan Duan, Rein Houthooft, John Schulman, Ilya Sutskever, and Pieter
  Abbeel.
\newblock Infogan: Interpretable representation learning by information
  maximizing generative adversarial nets.
\newblock In {\em NeurIPS}, 2016.

\bibitem{Vries17}
Harm de Vries, Florian Strub, Jeremie Mary, Hugo Larochelle, Olivier Pietquin,
  and Aaron Courville.
\newblock Modulating early visual processing by language.
\newblock In {\em NIPS}, 2017.

\bibitem{imagenet_cvpr09}
J. Deng, W. Dong, R. Socher, L.-J. Li, K. Li, and L. Fei-Fei.
\newblock {ImageNet: A Large-Scale Hierarchical Image Database}.
\newblock In {\em CVPR09}, 2009.

\bibitem{biggan}
Jeff Donahue, Philipp Krähenbühl, and Trevor Darrell.
\newblock Adversarial feature learning.
\newblock In {\em ICLR}, 2017.

\bibitem{bigbigan}
Jeff Donahue and Karen Simonyan.
\newblock Large scale adversarial representation learning.
\newblock In {\em NeurIPS}, 2019.

\bibitem{ali}
Vincent Dumoulin, Ishmael Belghazi, Ben Poole, Olivier Mastropietro, Alex Lamb,
  Martin Arjovsky, and Aaron Courville.
\newblock Adversarially learned inference.
\newblock In {\em ICLR}, 2017.

\bibitem{Dumoulin17}
Vincent Dumoulin, Jonathon Shlens, and Manjunath Kudlur.
\newblock A learned representation for artistic style.
\newblock In {\em ICLR}, 2017.

\bibitem{rotation}
Spyros Gidaris, Praveer Singh, and Nikos Komodakis.
\newblock Unsupervised representation learning by predicting image rotations.
\newblock In {\em ICLR}, 2018.

\bibitem{GANs}
Ian~J. Goodfellow, Jean Pouget-Abadie, Mehdi Mirza, Bing Xu, David
  Warde-Farley, Sherjil Ozair, Aaron Courville, and Yoshua Bengio.
\newblock Generative adversarial nets.
\newblock In {\em NeurIPS}, 2014.

\bibitem{ADC}
Philip Haeusser, Johannes Plapp, Vladimir Golkov, Elie Aljalbout, and Daniel
  Cremers.
\newblock Associative deep clustering: Training a classification network with
  no labels.
\newblock In {\em GCPR}, 2018.

\bibitem{resnet}
Kaiming He, Xiangyu Zhang, Shaoqing Ren, and Jian Sun.
\newblock Deep residual learning for image recognition, 2015.

\bibitem{FID}
Martin Heusel, Hubert Ramsauer, Thomas Unterthiner, Bernhard Nessler, and Sepp
  Hochreiter.
\newblock Gans trained by a two time-scale update rule converge to a local nash
  equilibrium.
\newblock In {\em NeurIPS}, 2017.

\bibitem{IMSAT}
Weihua Hu, Takeru Miyato, Seiya Tokui, Eiichi Matsumoto, and Masashi Sugiyama.
\newblock Learning discrete representations via information maximizing
  self-augmented training.
\newblock In {\em ICML}, 2017.

\bibitem{ji2019invariant}
Xu Ji, Jo{\~a}o~F Henriques, and Andrea Vedaldi.
\newblock Invariant information clustering for unsupervised image
  classification and segmentation.
\newblock In {\em ICCV}, 2019.

\bibitem{Adam}
Diederik~P Kingma and Jimmy~Ba. Adam.
\newblock A method for stochastic optimization.
\newblock In {\em ICLR}, 2015.

\bibitem{Hyun17}
Jae~Hyun Lim and Jong~Chul Ye.
\newblock Geometric gan.
\newblock In {\em arXiv}, 2017.

\bibitem{liu2020selfconditioned}
Steven Liu, Tongzhou Wang, David Bau, Jun-Yan Zhu, and Antonio Torralba.
\newblock Diverse image generation via self-conditioned gans.
\newblock In {\em CVPR}, 2020.

\bibitem{FewerLabels}
Mario Lucic, Michael Tschannen, Marvin Ritter, Xiaohua Zhai, Olivier Bachem,
  and Sylvain Gelly.
\newblock High-fidelity image generation with fewer labels.
\newblock In {\em ICML}, 2019.

\bibitem{cgan}
Mehdi Mirza and Simon Osindero.
\newblock Conditional generative adversarial nets.
\newblock In {\em NeurIPS}, 2014.

\bibitem{MiyatoProjection}
Takeru Miyato and Masanori Koyama.
\newblock cgans with projection discriminator.
\newblock In {\em ICLR}, 2018.

\bibitem{MiyatoSpectral}
Takeru Miyato and Masanori Koyama.
\newblock Spectral normalization for generative adversarial networks.
\newblock In {\em ICLR}, 2018.

\bibitem{clusterGAN}
Sudipto Mukherjee, Himanshu Asnani, Eugene Lin, and Sreeram Kannan.
\newblock Clustergan : Latent space clustering in generative adversarial
  networks.
\newblock In {\em AAAI}, 2019.

\bibitem{DCGAN}
Alec Radford, Luke Metz, and Soumith Chintala.
\newblock Unsupervised representation learning with deep convolutional
  generative adversarial network.
\newblock In {\em ICLR}, 2016.

\bibitem{insecption}
Tim Salimans, Ian Goodfellow, Wojciech Zaremba, Vicki Cheung, Alec Radford, and
  Xi Chen.
\newblock Improved techniques for training gans.
\newblock In {\em arXiv}, 2016.

\bibitem{infonce}
Kihyuk Sohn.
\newblock Improved deep metric learning with multi-class n-pair loss objective.
  in advances in neural information processing systems.
\newblock In {\em NIPS}, 2016.

\bibitem{Tran17}
Dustin Tran, Rajesh Ranganath, and David~M. Blei.
\newblock Hierarchical implicit models and likelihood-free variational
  inference.
\newblock In {\em NIPS}, 2017.

\bibitem{DEC}
Junyuan Xie, Ross Girshick, and Ali Farhadi.
\newblock Unsupervised deep embedding for clustering analysis.
\newblock In {\em ICML}, 2016.

\bibitem{lsun}
Fisher Yu, Yinda Zhang, Shuran Song, Ari Seff, and Jianxiong Xiao.
\newblock Lsun: Construction of a large-scale image dataset using deep learning
  with humans in the loop.
\newblock 2015.

\end{thebibliography}
}

\newpage
\section*{Appendix}

\begin{figure*}[htb]
\begin{minipage}[t]{.496\textwidth}
	\centering
	Generated cluster
\end{minipage}
\begin{minipage}[t]{.496\textwidth}
	\centering
	Real cluster
\end{minipage}
\vspace{.1cm}

\includegraphics[width = .5\textwidth]{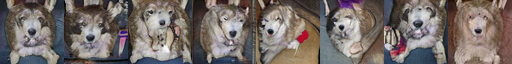}
\includegraphics[width = .5\textwidth]{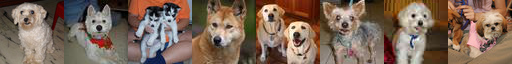}

\includegraphics[width = .5\textwidth]{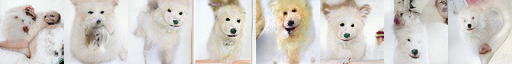}
\includegraphics[width = .5\textwidth]{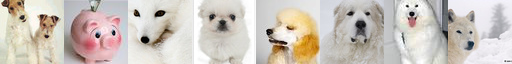}

\includegraphics[width = .5\textwidth]{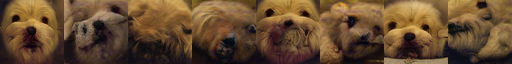}
\includegraphics[width = .5\textwidth]{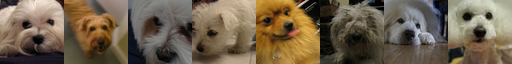}

\includegraphics[width = .5\textwidth]{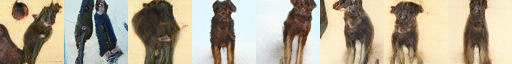}
\includegraphics[width = .5\textwidth]{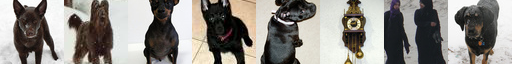}

\includegraphics[width = .5\textwidth]{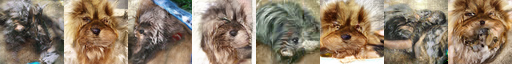}
\includegraphics[width = .5\textwidth]{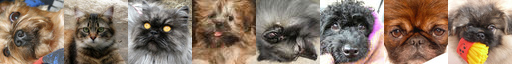}

\includegraphics[width = .5\textwidth]{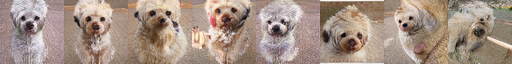}
\includegraphics[width = .5\textwidth]{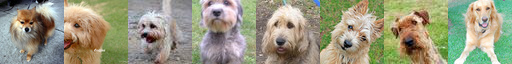}

\includegraphics[width = .5\textwidth]{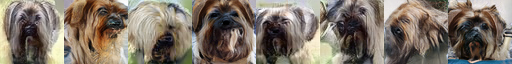}
\includegraphics[width = .5\textwidth]{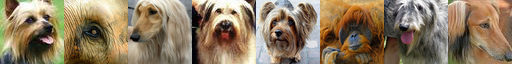}

\includegraphics[width = .5\textwidth]{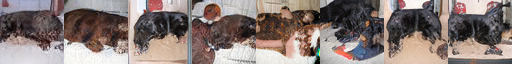}
\includegraphics[width = .5\textwidth]{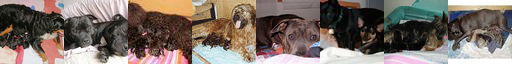}

\includegraphics[width = .5\textwidth]{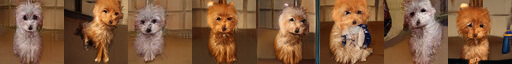}
\includegraphics[width = .5\textwidth]{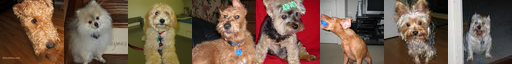}

\includegraphics[width = .5\textwidth]{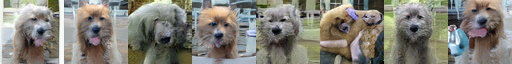}
\includegraphics[width = .5\textwidth]{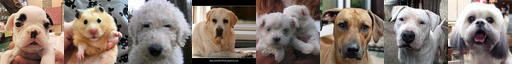}

\includegraphics[width = .5\textwidth]{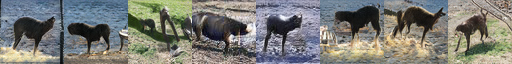}
\includegraphics[width = .5\textwidth]{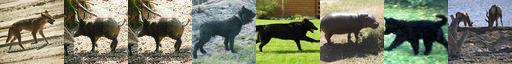}

\includegraphics[width = .5\textwidth]{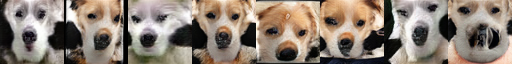}
\includegraphics[width = .5\textwidth]{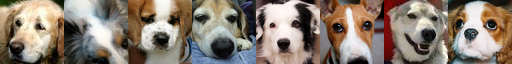}

\includegraphics[width = .5\textwidth]{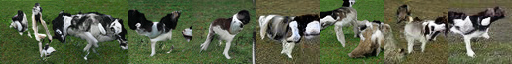}
\includegraphics[width = .5\textwidth]{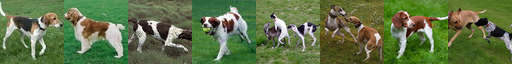}

\includegraphics[width = .5\textwidth]{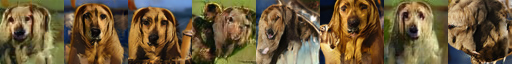}
\includegraphics[width = .5\textwidth]{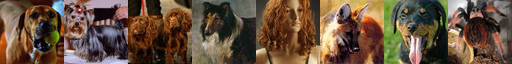}

\includegraphics[width = .5\textwidth]{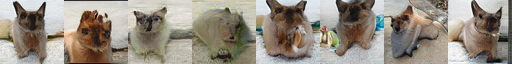}
\includegraphics[width = .5\textwidth]{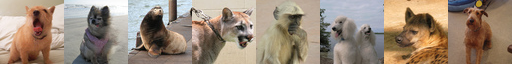}

\includegraphics[width = .5\textwidth]{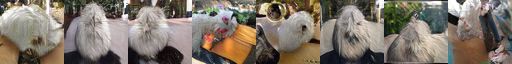}
\includegraphics[width = .5\textwidth]{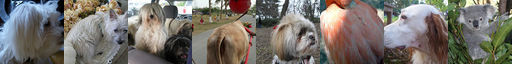}

\includegraphics[width = .5\textwidth]{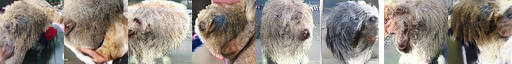}
\includegraphics[width = .5\textwidth]{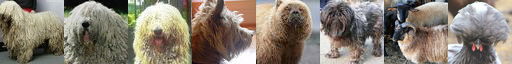}

\includegraphics[width = .5\textwidth]{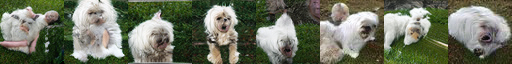}
\includegraphics[width = .5\textwidth]{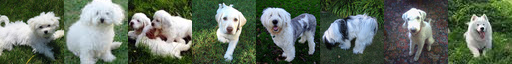}

\includegraphics[width = .5\textwidth]{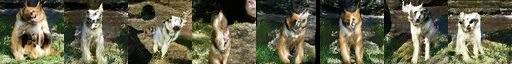}
\includegraphics[width = .5\textwidth]{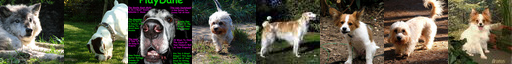}

\caption{Visualization of the generated and real images of several clusters for ImageNet. Each row shows $8$ samples with the highest confidence score by the clustering network of a cluster and the generated images conditioned on the same pseudo-label.}
\label{fig:fig1}

\end{figure*}

We perform more qualitative evaluations here. We visualize real and generated clusters without using any manual operation based on the confidence score of the clustering network. In the case of generated images, we generate $40$ samples and sort them based on the clustering network confidence score on the true cluster index, which is equal to the conditioning term by which the images are generated. Similarly, the real images in each cluster are sorted based on the confidence scores of the clustering network. We show $8$  real and generated samples with the highest confidence scores for $171$ clusters for ImageNet in Figure~\ref{fig:fig1}. Our model is trained with $1000$ clusters.

\begin{figure*}[htb]
\begin{minipage}[t]{.496\textwidth}
	\centering
	Generated cluster
\end{minipage}
\begin{minipage}[t]{.496\textwidth}
	\centering
	Real cluster
\end{minipage}
\vspace{.1cm}

\includegraphics[width = .5\textwidth]{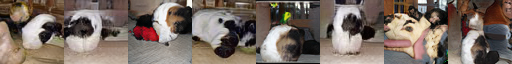}
\includegraphics[width = .5\textwidth]{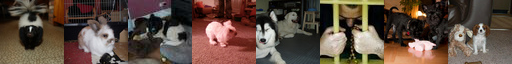}

\includegraphics[width = .5\textwidth]{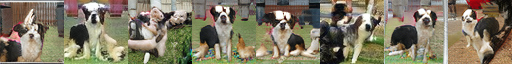}
\includegraphics[width = .5\textwidth]{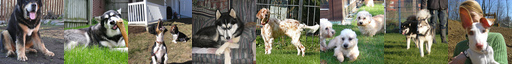}

\includegraphics[width = .5\textwidth]{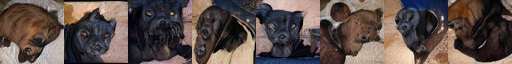}
\includegraphics[width = .5\textwidth]{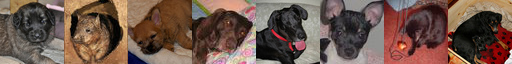}

\includegraphics[width = .5\textwidth]{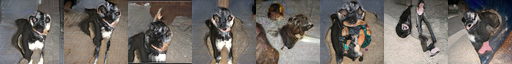}
\includegraphics[width = .5\textwidth]{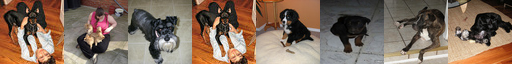}

\includegraphics[width = .5\textwidth]{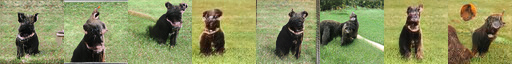}
\includegraphics[width = .5\textwidth]{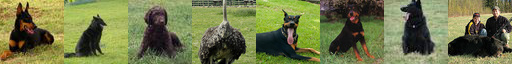}

\includegraphics[width = .5\textwidth]{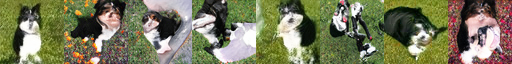}
\includegraphics[width = .5\textwidth]{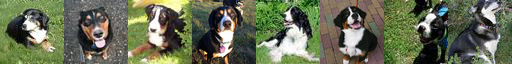}

\includegraphics[width = .5\textwidth]{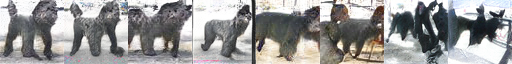}
\includegraphics[width = .5\textwidth]{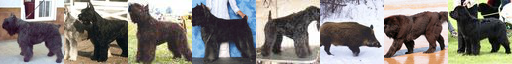}

\includegraphics[width = .5\textwidth]{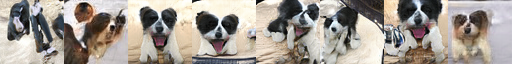}
\includegraphics[width = .5\textwidth]{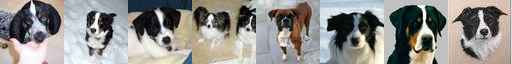}

\includegraphics[width = .5\textwidth]{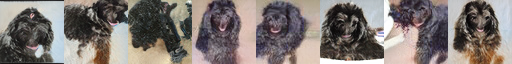}
\includegraphics[width = .5\textwidth]{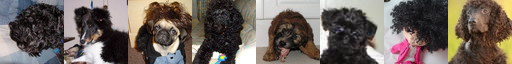}

\includegraphics[width = .5\textwidth]{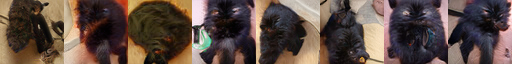}
\includegraphics[width = .5\textwidth]{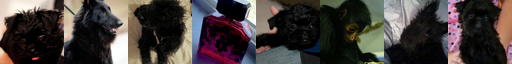}

\includegraphics[width = .5\textwidth]{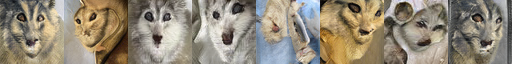}
\includegraphics[width = .5\textwidth]{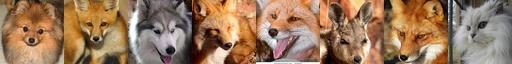}

\includegraphics[width = .5\textwidth]{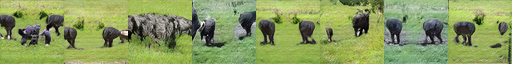}
\includegraphics[width = .5\textwidth]{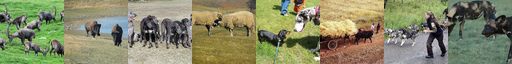}

\includegraphics[width = .5\textwidth]{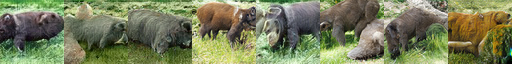}
\includegraphics[width = .5\textwidth]{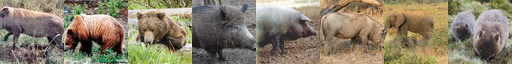}

\includegraphics[width = .5\textwidth]{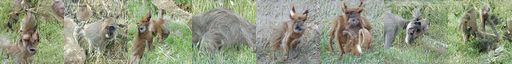}
\includegraphics[width = .5\textwidth]{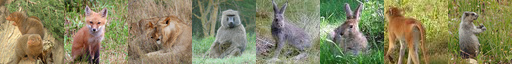}

\includegraphics[width = .5\textwidth]{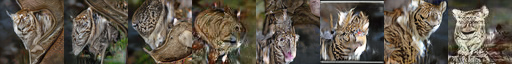}
\includegraphics[width = .5\textwidth]{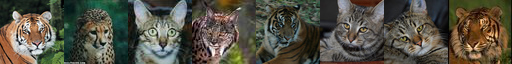}

\includegraphics[width = .5\textwidth]{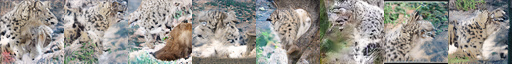}
\includegraphics[width = .5\textwidth]{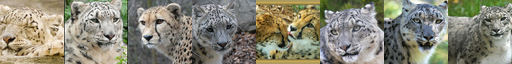}

\includegraphics[width = .5\textwidth]{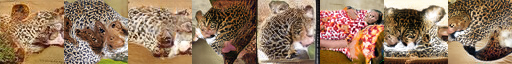}
\includegraphics[width = .5\textwidth]{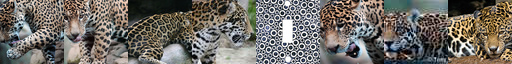}

\includegraphics[width = .5\textwidth]{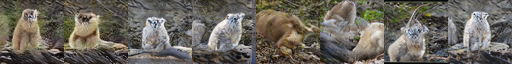}
\includegraphics[width = .5\textwidth]{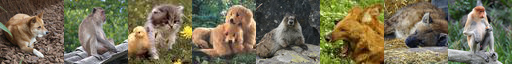}

\includegraphics[width = .5\textwidth]{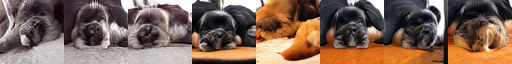}
\includegraphics[width = .5\textwidth]{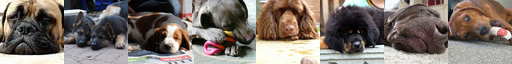}

\begin{minipage}[t]{1\textwidth}
Figure 1: Visualization of the generated and real images of several clusters for ImageNet. Each row shows $8$ samples with the highest confidence score by the clustering network of a cluster and the generated images conditioned on the same pseudo-label.
\end{minipage}

\end{figure*}

\begin{figure*}[htb]
\begin{minipage}[t]{.496\textwidth}
	\centering
	Generated cluster
\end{minipage}
\begin{minipage}[t]{.496\textwidth}
	\centering
	Real cluster
\end{minipage}
\vspace{.1cm}

\includegraphics[width = .5\textwidth]{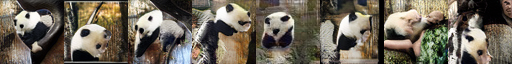}
\includegraphics[width = .5\textwidth]{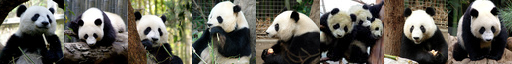}

\includegraphics[width = .5\textwidth]{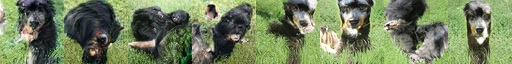}
\includegraphics[width = .5\textwidth]{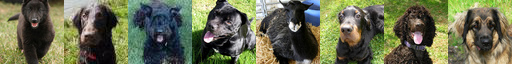}

\includegraphics[width = .5\textwidth]{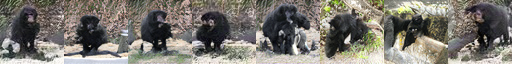}
\includegraphics[width = .5\textwidth]{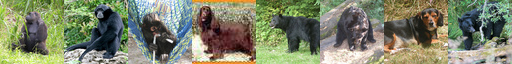}

\includegraphics[width = .5\textwidth]{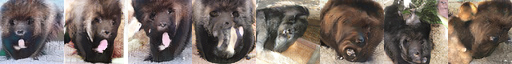}
\includegraphics[width = .5\textwidth]{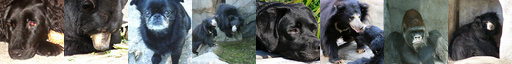}

\includegraphics[width = .5\textwidth]{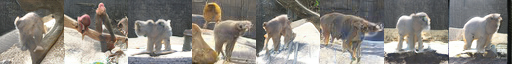}
\includegraphics[width = .5\textwidth]{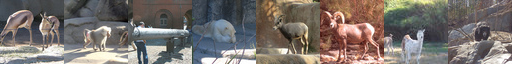}

\includegraphics[width = .5\textwidth]{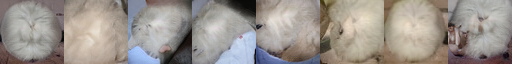}
\includegraphics[width = .5\textwidth]{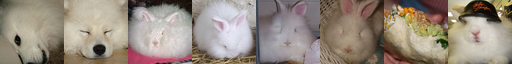}

\includegraphics[width = .5\textwidth]{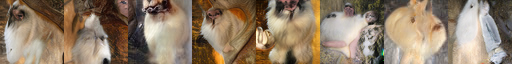}
\includegraphics[width = .5\textwidth]{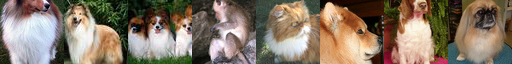}

\includegraphics[width = .5\textwidth]{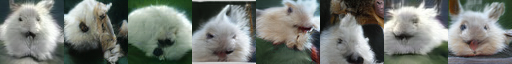}
\includegraphics[width = .5\textwidth]{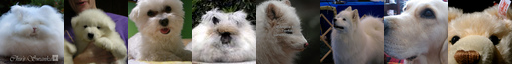}

\includegraphics[width = .5\textwidth]{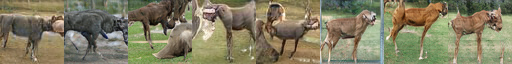}
\includegraphics[width = .5\textwidth]{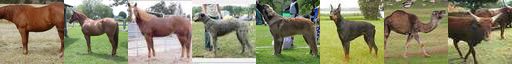}

\includegraphics[width = .5\textwidth]{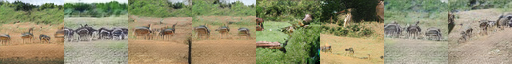}
\includegraphics[width = .5\textwidth]{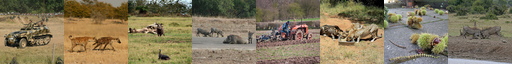}

\includegraphics[width = .5\textwidth]{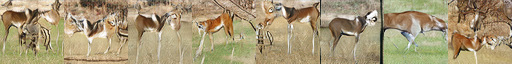}
\includegraphics[width = .5\textwidth]{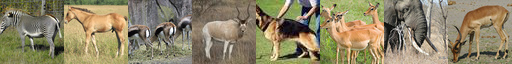}

\includegraphics[width = .5\textwidth]{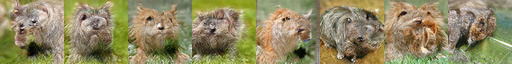}
\includegraphics[width = .5\textwidth]{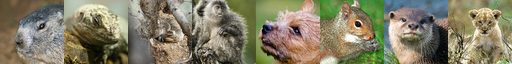}

\includegraphics[width = .5\textwidth]{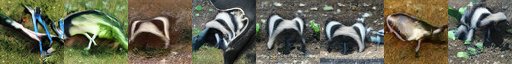}
\includegraphics[width = .5\textwidth]{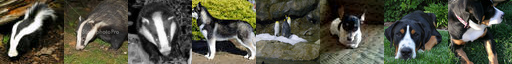}

\includegraphics[width = .5\textwidth]{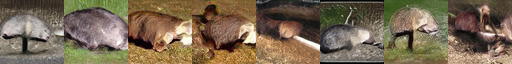}
\includegraphics[width = .5\textwidth]{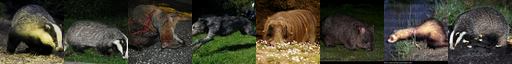}

\includegraphics[width = .5\textwidth]{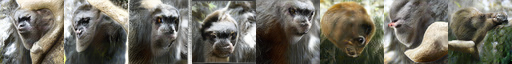}
\includegraphics[width = .5\textwidth]{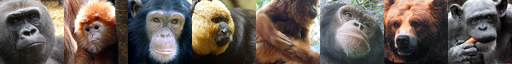}

\includegraphics[width = .5\textwidth]{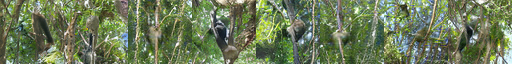}
\includegraphics[width = .5\textwidth]{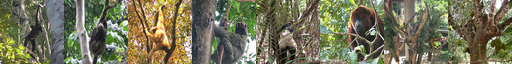}

\includegraphics[width = .5\textwidth]{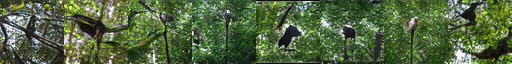}
\includegraphics[width = .5\textwidth]{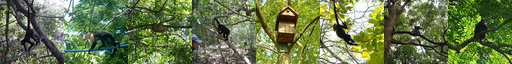}

\includegraphics[width = .5\textwidth]{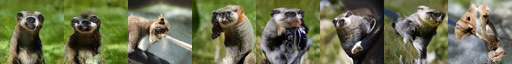}
\includegraphics[width = .5\textwidth]{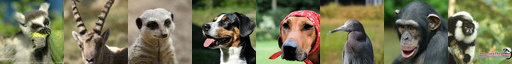}

\includegraphics[width = .5\textwidth]{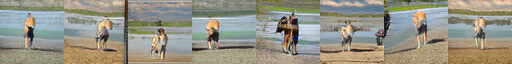}
\includegraphics[width = .5\textwidth]{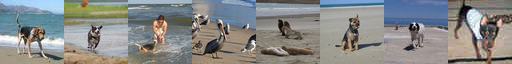}

\begin{minipage}[t]{1\textwidth}
Figure 1: Visualization of the generated and real images of several clusters for ImageNet. Each row shows $8$ samples with the highest confidence score by the clustering network of a cluster and the generated images conditioned on the same pseudo-label.
\end{minipage}

\end{figure*}

\begin{figure*}[htb]
\begin{minipage}[t]{.496\textwidth}
	\centering
	Generated cluster
\end{minipage}
\begin{minipage}[t]{.496\textwidth}
	\centering
	Real cluster
\end{minipage}
\vspace{.1cm}

\includegraphics[width = .5\textwidth]{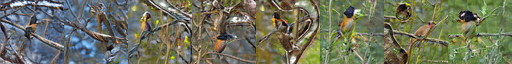}
\includegraphics[width = .5\textwidth]{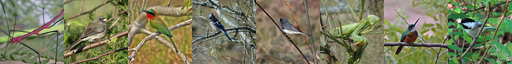}

\includegraphics[width = .5\textwidth]{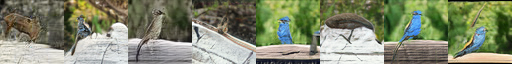}
\includegraphics[width = .5\textwidth]{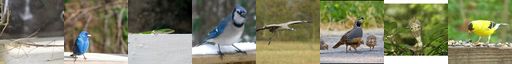}

\includegraphics[width = .5\textwidth]{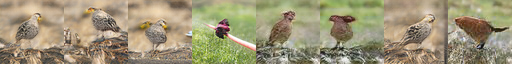}
\includegraphics[width = .5\textwidth]{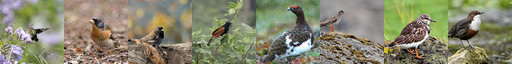}

\includegraphics[width = .5\textwidth]{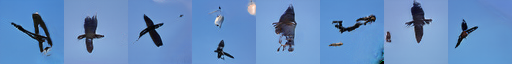}
\includegraphics[width = .5\textwidth]{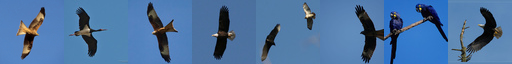}

\includegraphics[width = .5\textwidth]{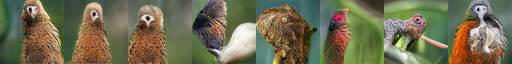}
\includegraphics[width = .5\textwidth]{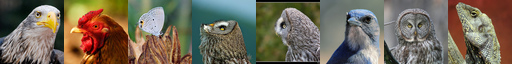}

\includegraphics[width = .5\textwidth]{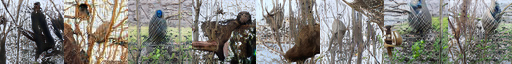}
\includegraphics[width = .5\textwidth]{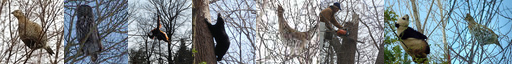}

\includegraphics[width = .5\textwidth]{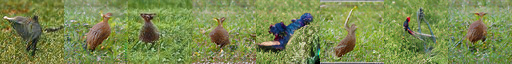}
\includegraphics[width = .5\textwidth]{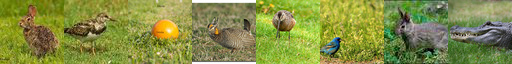}

\includegraphics[width = .5\textwidth]{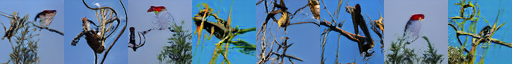}
\includegraphics[width = .5\textwidth]{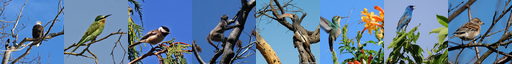}

\includegraphics[width = .5\textwidth]{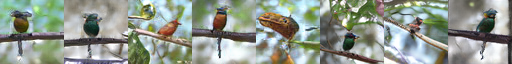}
\includegraphics[width = .5\textwidth]{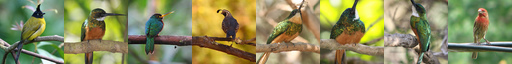}

\includegraphics[width = .5\textwidth]{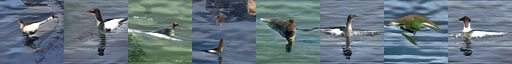}
\includegraphics[width = .5\textwidth]{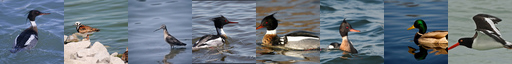}

\includegraphics[width = .5\textwidth]{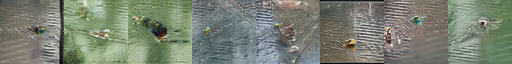}
\includegraphics[width = .5\textwidth]{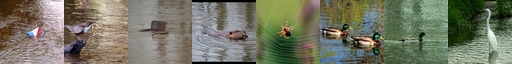}

\includegraphics[width = .5\textwidth]{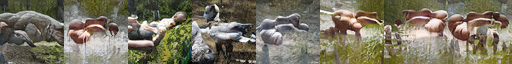}
\includegraphics[width = .5\textwidth]{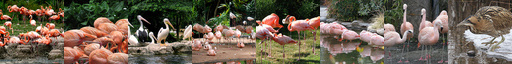}

\includegraphics[width = .5\textwidth]{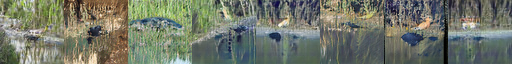}
\includegraphics[width = .5\textwidth]{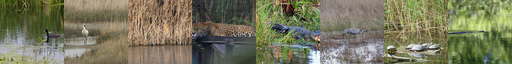}

\includegraphics[width = .5\textwidth]{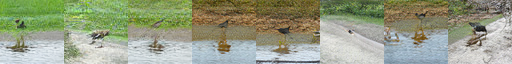}
\includegraphics[width = .5\textwidth]{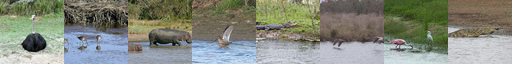}

\includegraphics[width = .5\textwidth]{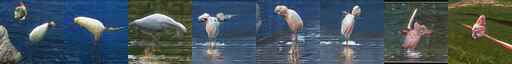}
\includegraphics[width = .5\textwidth]{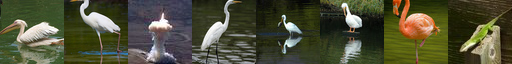}

\includegraphics[width = .5\textwidth]{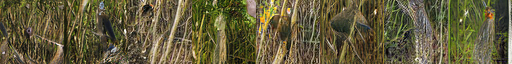}
\includegraphics[width = .5\textwidth]{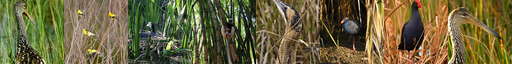}

\includegraphics[width = .5\textwidth]{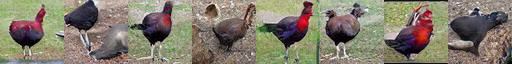}
\includegraphics[width = .5\textwidth]{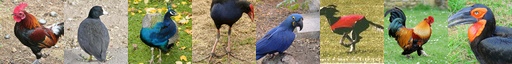}

\includegraphics[width = .5\textwidth]{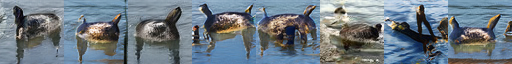}
\includegraphics[width = .5\textwidth]{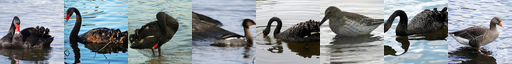}

\includegraphics[width = .5\textwidth]{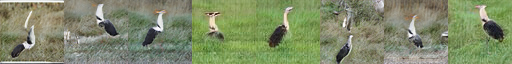}
\includegraphics[width = .5\textwidth]{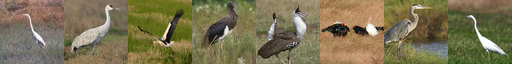}

\begin{minipage}[t]{1\textwidth}
Figure 1: Visualization of the generated and real images of several clusters for ImageNet. Each row shows $8$ samples with the highest confidence score by the clustering network of a cluster and the generated images conditioned on the same pseudo-label.
\end{minipage}

\end{figure*}

\begin{figure*}[htb]
\begin{minipage}[t]{.496\textwidth}
	\centering
	Generated cluster
\end{minipage}
\begin{minipage}[t]{.496\textwidth}
	\centering
	Real cluster
\end{minipage}
\vspace{.1cm}

\includegraphics[width = .5\textwidth]{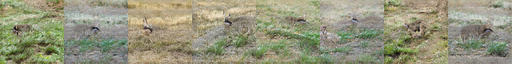}
\includegraphics[width = .5\textwidth]{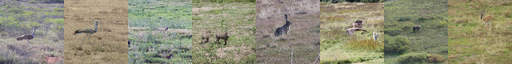}

\includegraphics[width = .5\textwidth]{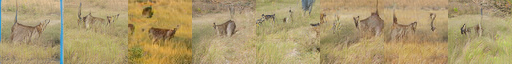}
\includegraphics[width = .5\textwidth]{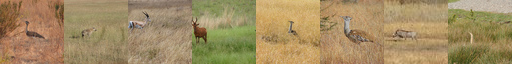}

\includegraphics[width = .5\textwidth]{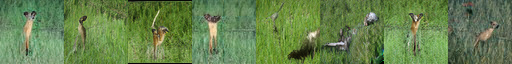}
\includegraphics[width = .5\textwidth]{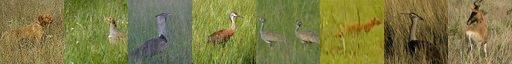}

\includegraphics[width = .5\textwidth]{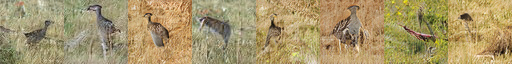}
\includegraphics[width = .5\textwidth]{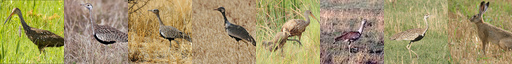}

\includegraphics[width = .5\textwidth]{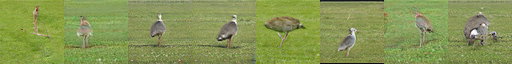}
\includegraphics[width = .5\textwidth]{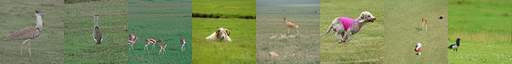}

\includegraphics[width = .5\textwidth]{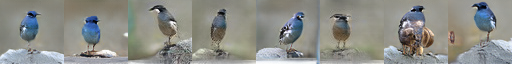}
\includegraphics[width = .5\textwidth]{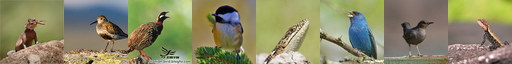}

\includegraphics[width = .5\textwidth]{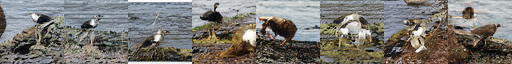}
\includegraphics[width = .5\textwidth]{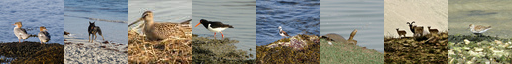}

\includegraphics[width = .5\textwidth]{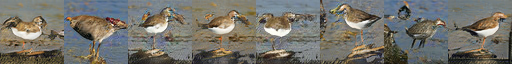}
\includegraphics[width = .5\textwidth]{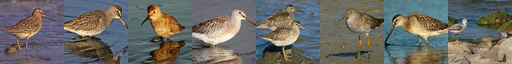}

\includegraphics[width = .5\textwidth]{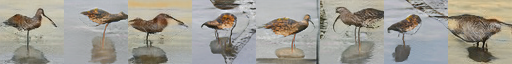}
\includegraphics[width = .5\textwidth]{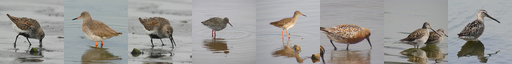}

\includegraphics[width = .5\textwidth]{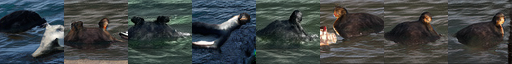}
\includegraphics[width = .5\textwidth]{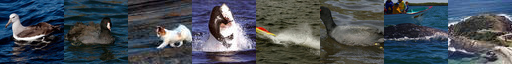}

\includegraphics[width = .5\textwidth]{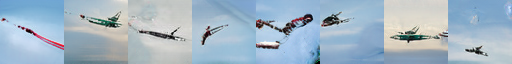}
\includegraphics[width = .5\textwidth]{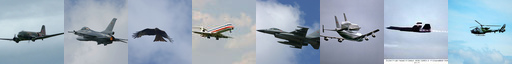}

\includegraphics[width = .5\textwidth]{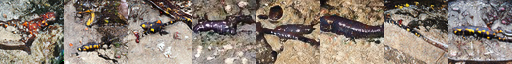}
\includegraphics[width = .5\textwidth]{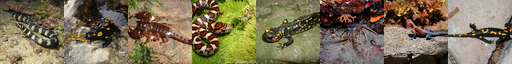}

\includegraphics[width = .5\textwidth]{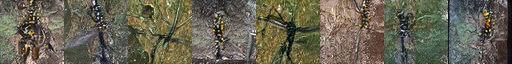}
\includegraphics[width = .5\textwidth]{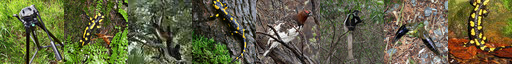}

\includegraphics[width = .5\textwidth]{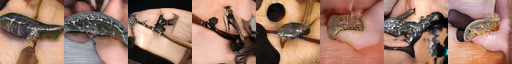}
\includegraphics[width = .5\textwidth]{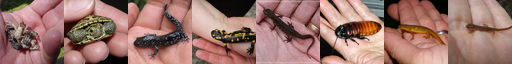}

\includegraphics[width = .5\textwidth]{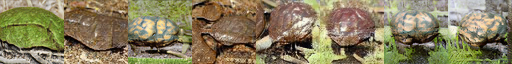}
\includegraphics[width = .5\textwidth]{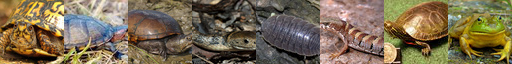}

\includegraphics[width = .5\textwidth]{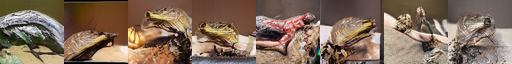}
\includegraphics[width = .5\textwidth]{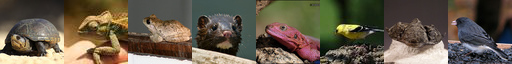}

\includegraphics[width = .5\textwidth]{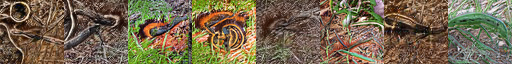}
\includegraphics[width = .5\textwidth]{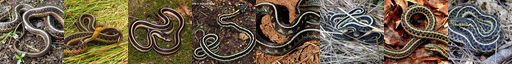}

\includegraphics[width = .5\textwidth]{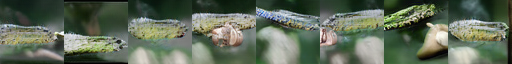}
\includegraphics[width = .5\textwidth]{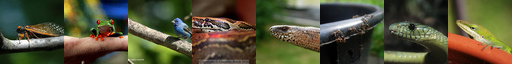}

\includegraphics[width = .5\textwidth]{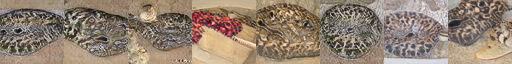}
\includegraphics[width = .5\textwidth]{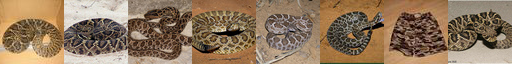}

\begin{minipage}[t]{1\textwidth}
Figure 1: Visualization of the generated and real images of several clusters for ImageNet. Each row shows $8$ samples with the highest confidence score by the clustering network of a cluster and the generated images conditioned on the same pseudo-label.
\end{minipage}

\end{figure*}

\begin{figure*}[htb]
\begin{minipage}[t]{.496\textwidth}
	\centering
	Generated cluster
\end{minipage}
\begin{minipage}[t]{.496\textwidth}
	\centering
	Real cluster
\end{minipage}
\vspace{.1cm}

\includegraphics[width = .5\textwidth]{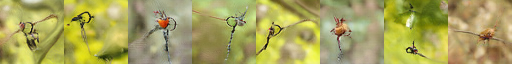}
\includegraphics[width = .5\textwidth]{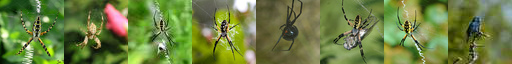}

\includegraphics[width = .5\textwidth]{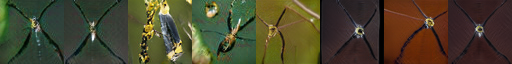}
\includegraphics[width = .5\textwidth]{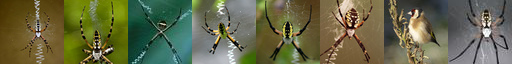}

\includegraphics[width = .5\textwidth]{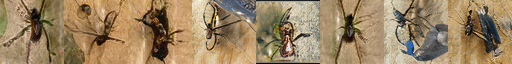}
\includegraphics[width = .5\textwidth]{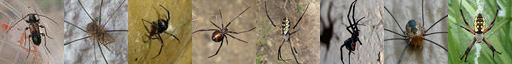}

\includegraphics[width = .5\textwidth]{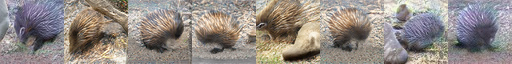}
\includegraphics[width = .5\textwidth]{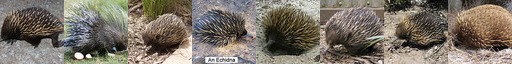}

\includegraphics[width = .5\textwidth]{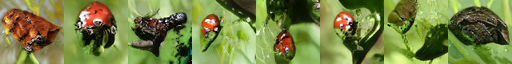}
\includegraphics[width = .5\textwidth]{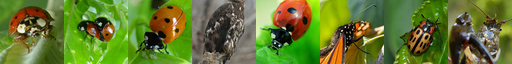}

\includegraphics[width = .5\textwidth]{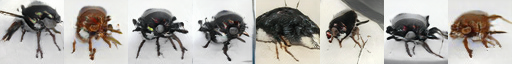}
\includegraphics[width = .5\textwidth]{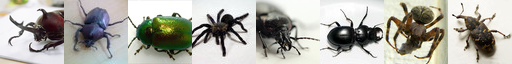}

\includegraphics[width = .5\textwidth]{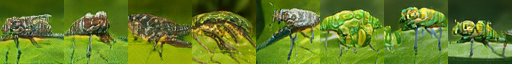}
\includegraphics[width = .5\textwidth]{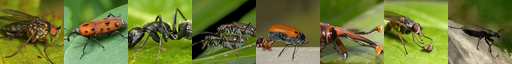}

\includegraphics[width = .5\textwidth]{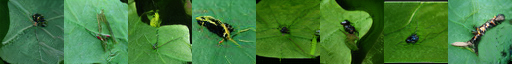}
\includegraphics[width = .5\textwidth]{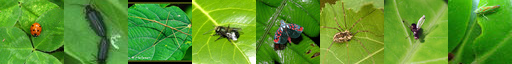}

\includegraphics[width = .5\textwidth]{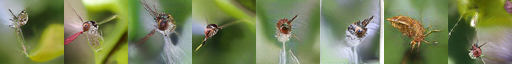}
\includegraphics[width = .5\textwidth]{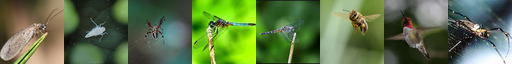}

\includegraphics[width = .5\textwidth]{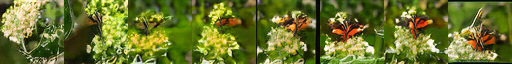}
\includegraphics[width = .5\textwidth]{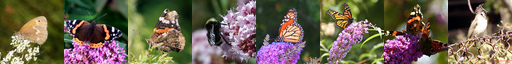}

\includegraphics[width = .5\textwidth]{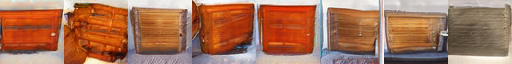}
\includegraphics[width = .5\textwidth]{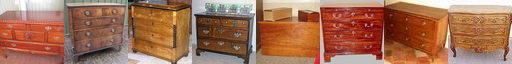}

\includegraphics[width = .5\textwidth]{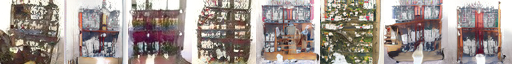}
\includegraphics[width = .5\textwidth]{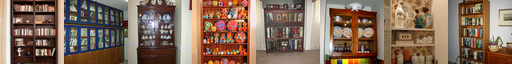}

\includegraphics[width = .5\textwidth]{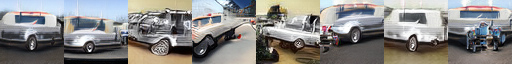}
\includegraphics[width = .5\textwidth]{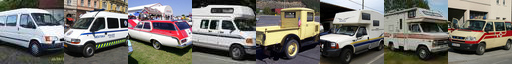}

\includegraphics[width = .5\textwidth]{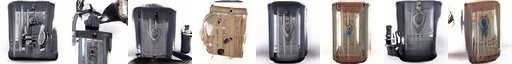}
\includegraphics[width = .5\textwidth]{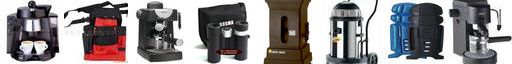}

\includegraphics[width = .5\textwidth]{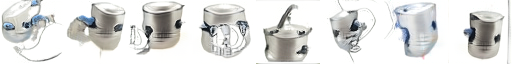}
\includegraphics[width = .5\textwidth]{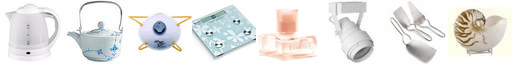}

\includegraphics[width = .5\textwidth]{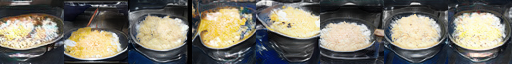}
\includegraphics[width = .5\textwidth]{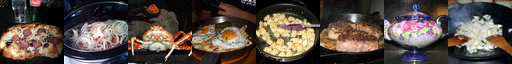}

\includegraphics[width = .5\textwidth]{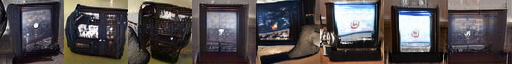}
\includegraphics[width = .5\textwidth]{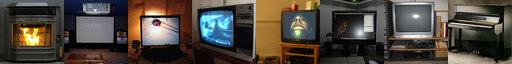}

\includegraphics[width = .5\textwidth]{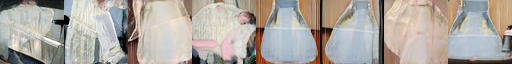}
\includegraphics[width = .5\textwidth]{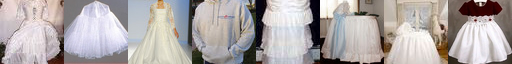}

\includegraphics[width = .5\textwidth]{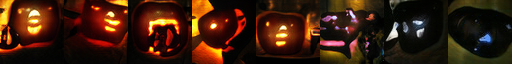}
\includegraphics[width = .5\textwidth]{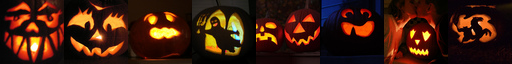}

\begin{minipage}[t]{1\textwidth}
Figure 1: Visualization of the generated and real images of several clusters for ImageNet. Each row shows $8$ samples with the highest confidence score by the clustering network of a cluster and the generated images conditioned on the same pseudo-label.
\end{minipage}

\end{figure*}

\begin{figure*}[htb]
\begin{minipage}[t]{.496\textwidth}
	\centering
	Generated cluster
\end{minipage}
\begin{minipage}[t]{.496\textwidth}
	\centering
	Real cluster
\end{minipage}
\vspace{.1cm}

\includegraphics[width = .5\textwidth]{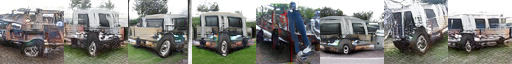}
\includegraphics[width = .5\textwidth]{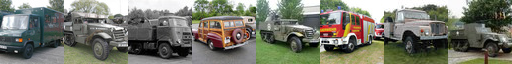}

\includegraphics[width = .5\textwidth]{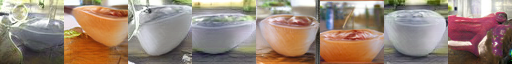}
\includegraphics[width = .5\textwidth]{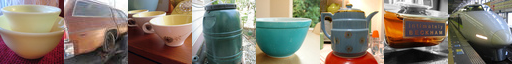}

\includegraphics[width = .5\textwidth]{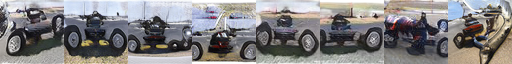}
\includegraphics[width = .5\textwidth]{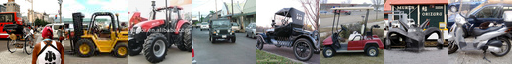}

\includegraphics[width = .5\textwidth]{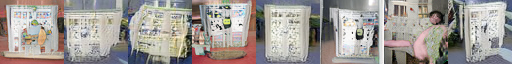}
\includegraphics[width = .5\textwidth]{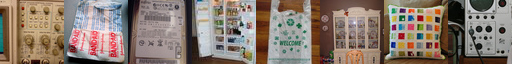}

\includegraphics[width = .5\textwidth]{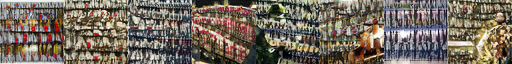}
\includegraphics[width = .5\textwidth]{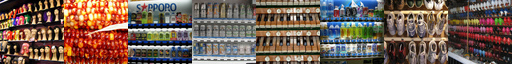}

\includegraphics[width = .5\textwidth]{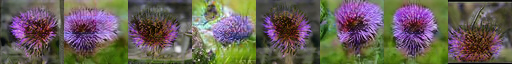}
\includegraphics[width = .5\textwidth]{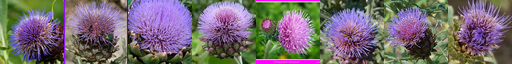}

\includegraphics[width = .5\textwidth]{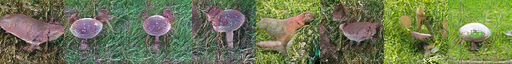}
\includegraphics[width = .5\textwidth]{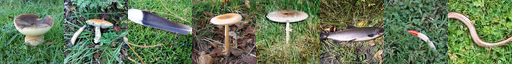}

\includegraphics[width = .5\textwidth]{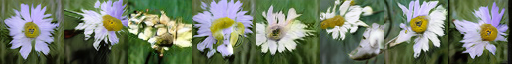}
\includegraphics[width = .5\textwidth]{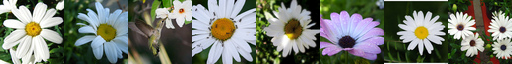}

\includegraphics[width = .5\textwidth]{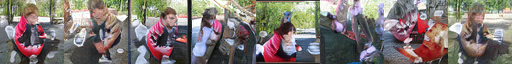}
\includegraphics[width = .5\textwidth]{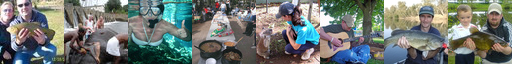}

\includegraphics[width = .5\textwidth]{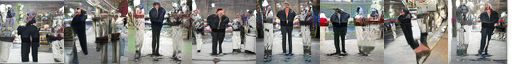}
\includegraphics[width = .5\textwidth]{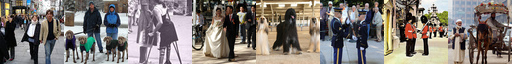}

\includegraphics[width = .5\textwidth]{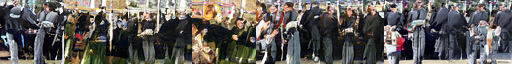}
\includegraphics[width = .5\textwidth]{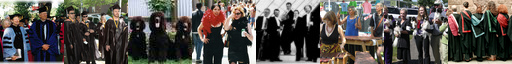}

\includegraphics[width = .5\textwidth]{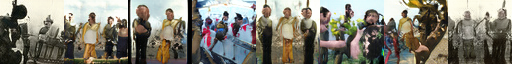}
\includegraphics[width = .5\textwidth]{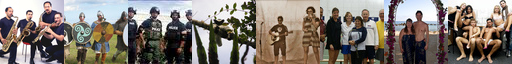}

\includegraphics[width = .5\textwidth]{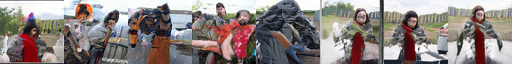}
\includegraphics[width = .5\textwidth]{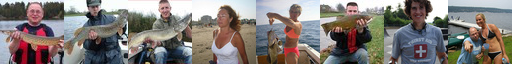}

\includegraphics[width = .5\textwidth]{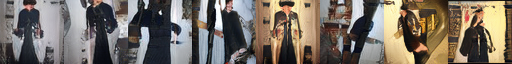}
\includegraphics[width = .5\textwidth]{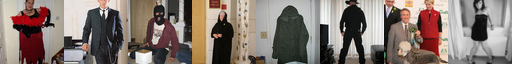}

\includegraphics[width = .5\textwidth]{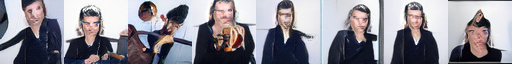}
\includegraphics[width = .5\textwidth]{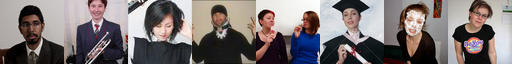}

\includegraphics[width = .5\textwidth]{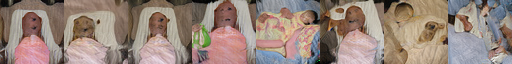}
\includegraphics[width = .5\textwidth]{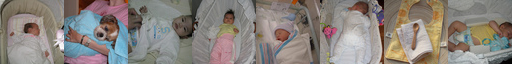}

\includegraphics[width = .5\textwidth]{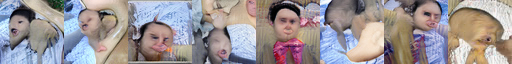}
\includegraphics[width = .5\textwidth]{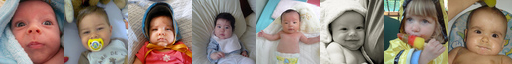}

\includegraphics[width = .5\textwidth]{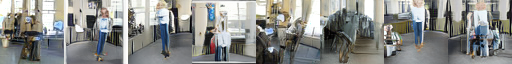}
\includegraphics[width = .5\textwidth]{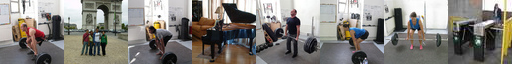}

\includegraphics[width = .5\textwidth]{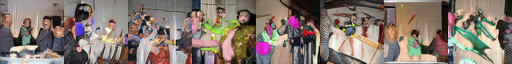}
\includegraphics[width = .5\textwidth]{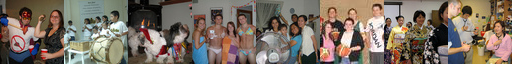}

\begin{minipage}[t]{1\textwidth}
Figure 1: Visualization of the generated and real images of several clusters for ImageNet. Each row shows $8$ samples with the highest confidence score by the clustering network of a cluster and the generated images conditioned on the same pseudo-label.
\end{minipage}

\end{figure*}

\begin{figure*}[htb]
\begin{minipage}[t]{.496\textwidth}
	\centering
	Generated cluster
\end{minipage}
\begin{minipage}[t]{.496\textwidth}
	\centering
	Real cluster
\end{minipage}

\vspace{.1cm}

\includegraphics[width = .5\textwidth]{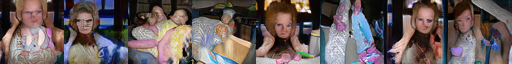}
\includegraphics[width = .5\textwidth]{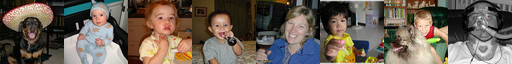}

\includegraphics[width = .5\textwidth]{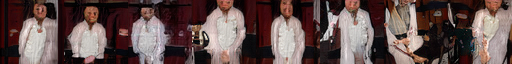}
\includegraphics[width = .5\textwidth]{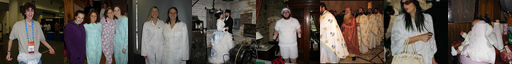}

\includegraphics[width = .5\textwidth]{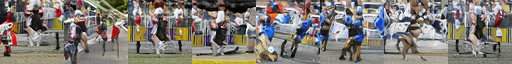}
\includegraphics[width = .5\textwidth]{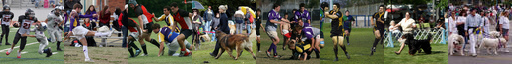}

\includegraphics[width = .5\textwidth]{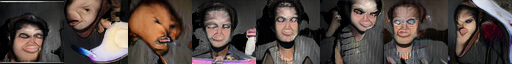}
\includegraphics[width = .5\textwidth]{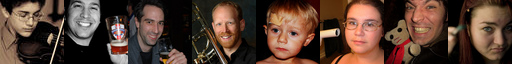}

\includegraphics[width = .5\textwidth]{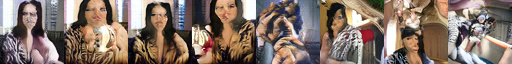}
\includegraphics[width = .5\textwidth]{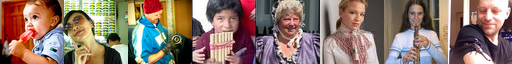}

\includegraphics[width = .5\textwidth]{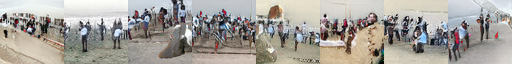}
\includegraphics[width = .5\textwidth]{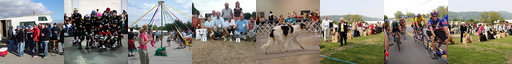}

\includegraphics[width = .5\textwidth]{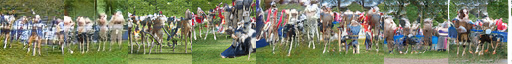}
\includegraphics[width = .5\textwidth]{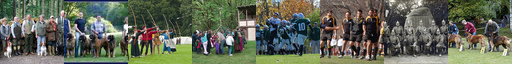}

\includegraphics[width = .5\textwidth]{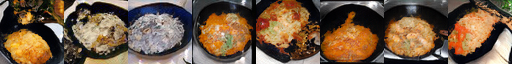}
\includegraphics[width = .5\textwidth]{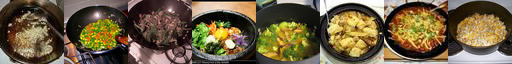}

\includegraphics[width = .5\textwidth]{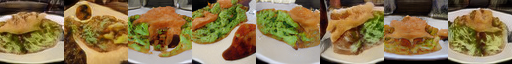}
\includegraphics[width = .5\textwidth]{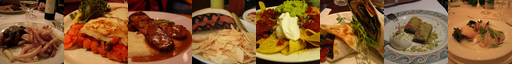}

\includegraphics[width = .5\textwidth]{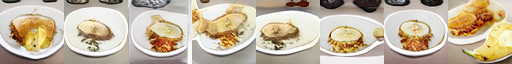}
\includegraphics[width = .5\textwidth]{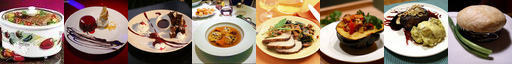}

\includegraphics[width = .5\textwidth]{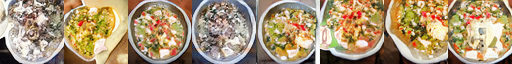}
\includegraphics[width = .5\textwidth]{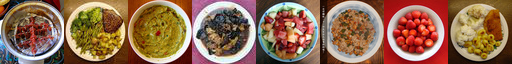}

\includegraphics[width = .5\textwidth]{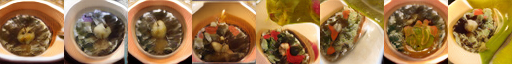}
\includegraphics[width = .5\textwidth]{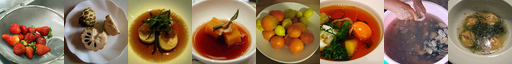}

\includegraphics[width = .5\textwidth]{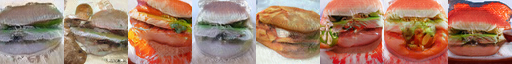}
\includegraphics[width = .5\textwidth]{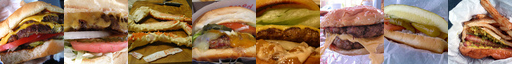}

\includegraphics[width = .5\textwidth]{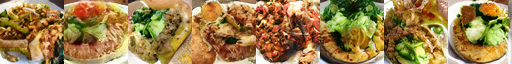}
\includegraphics[width = .5\textwidth]{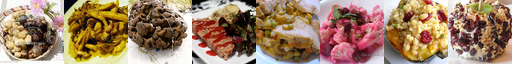}

\includegraphics[width = .5\textwidth]{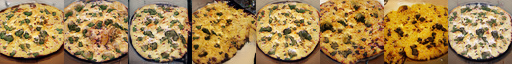}
\includegraphics[width = .5\textwidth]{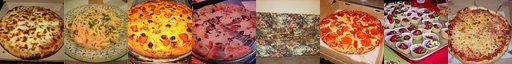}

\includegraphics[width = .5\textwidth]{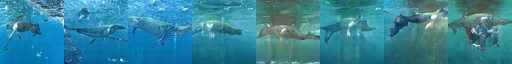}
\includegraphics[width = .5\textwidth]{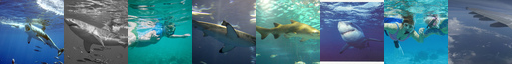}

\includegraphics[width = .5\textwidth]{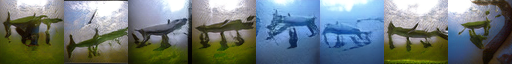}
\includegraphics[width = .5\textwidth]{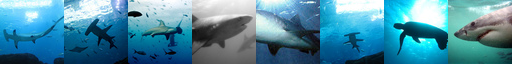}

\includegraphics[width = .5\textwidth]{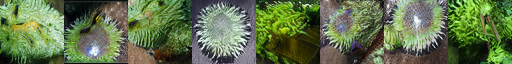}
\includegraphics[width = .5\textwidth]{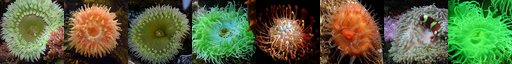}

\includegraphics[width = .5\textwidth]{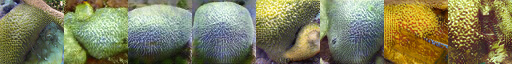}
\includegraphics[width = .5\textwidth]{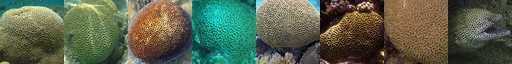}

\begin{minipage}[t]{1\textwidth}
Figure 1: Visualization of the generated and real images of several clusters for ImageNet. Each row shows $8$ samples with the highest confidence score by the clustering network of a cluster and the generated images conditioned on the same pseudo-label.
\end{minipage}

\end{figure*}

\begin{figure*}[htb]
\begin{minipage}[t]{.496\textwidth}
	\centering
	Generated cluster
\end{minipage}
\begin{minipage}[t]{.496\textwidth}
	\centering
	Real cluster
\end{minipage}

\vspace{.1cm}

\includegraphics[width = .5\textwidth]{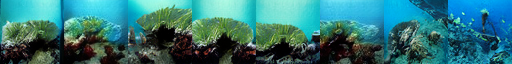}
\includegraphics[width = .5\textwidth]{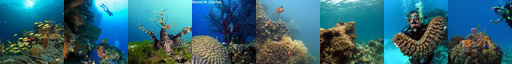}

\includegraphics[width = .5\textwidth]{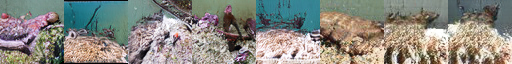}
\includegraphics[width = .5\textwidth]{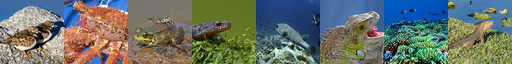}

\includegraphics[width = .5\textwidth]{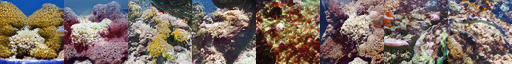}
\includegraphics[width = .5\textwidth]{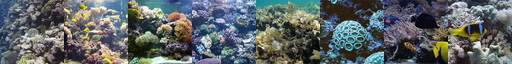}

\includegraphics[width = .5\textwidth]{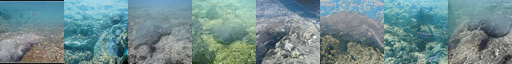}
\includegraphics[width = .5\textwidth]{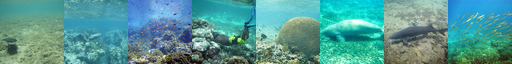}

\includegraphics[width = .5\textwidth]{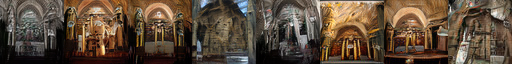}
\includegraphics[width = .5\textwidth]{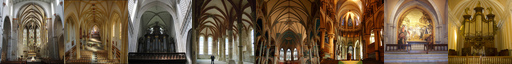}

\includegraphics[width = .5\textwidth]{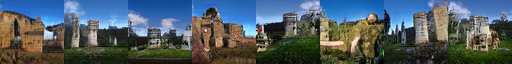}
\includegraphics[width = .5\textwidth]{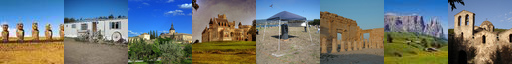}

\includegraphics[width = .5\textwidth]{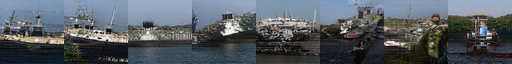}
\includegraphics[width = .5\textwidth]{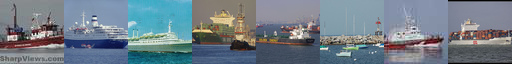}

\includegraphics[width = .5\textwidth]{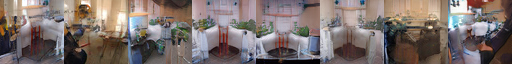}
\includegraphics[width = .5\textwidth]{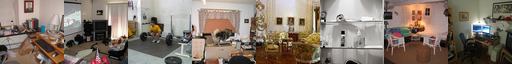}

\includegraphics[width = .5\textwidth]{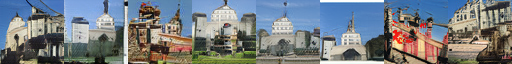}
\includegraphics[width = .5\textwidth]{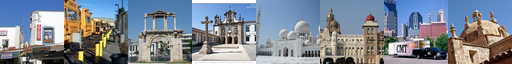}

\includegraphics[width = .5\textwidth]{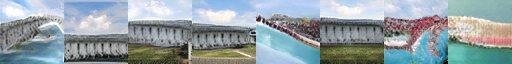}
\includegraphics[width = .5\textwidth]{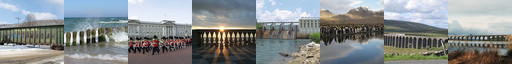}

\includegraphics[width = .5\textwidth]{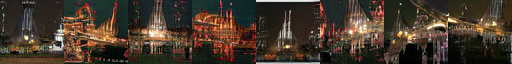}
\includegraphics[width = .5\textwidth]{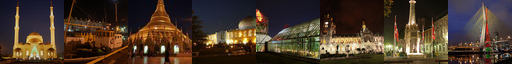}

\includegraphics[width = .5\textwidth]{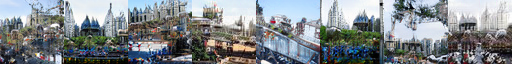}
\includegraphics[width = .5\textwidth]{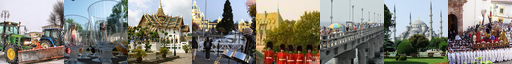}

\includegraphics[width = .5\textwidth]{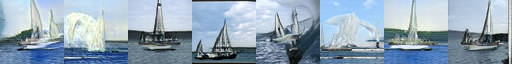}
\includegraphics[width = .5\textwidth]{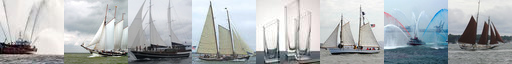}

\includegraphics[width = .5\textwidth]{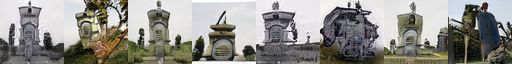}
\includegraphics[width = .5\textwidth]{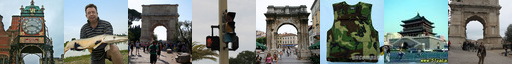}

\includegraphics[width = .5\textwidth]{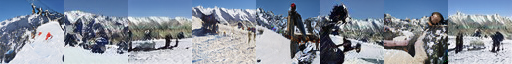}
\includegraphics[width = .5\textwidth]{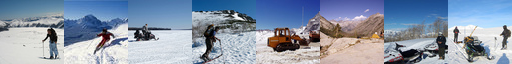}

\includegraphics[width = .5\textwidth]{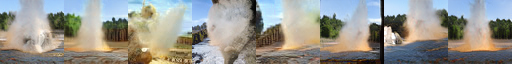}
\includegraphics[width = .5\textwidth]{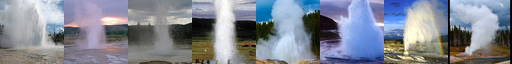}

\includegraphics[width = .5\textwidth]{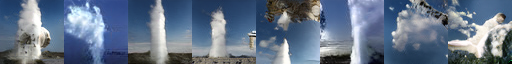}
\includegraphics[width = .5\textwidth]{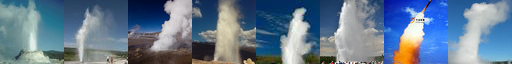}

\includegraphics[width = .5\textwidth]{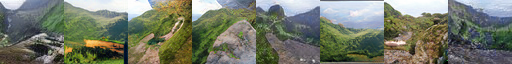}
\includegraphics[width = .5\textwidth]{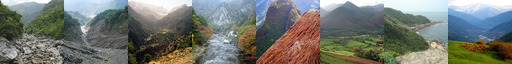}

\includegraphics[width = .5\textwidth]{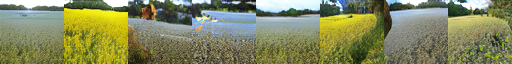}
\includegraphics[width = .5\textwidth]{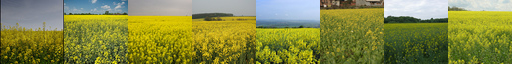}

\begin{minipage}[t]{1\textwidth}
Figure 1: Visualization of the generated and real images of several clusters for ImageNet. Each row shows $8$ samples with the highest confidence score by the clustering network of a cluster and the generated images conditioned on the same pseudo-label.
\end{minipage}

\end{figure*}

\end{document}